\documentclass{article} 
\usepackage{iclr2027_conference,times}

\usepackage{amsmath,amsfonts,bm}

\def\eqref#1{equation~\ref{#1}}

\def\1{\bm{1}}

\DeclareMathAlphabet{\mathsfit}{\encodingdefault}{\sfdefault}{m}{sl}
\SetMathAlphabet{\mathsfit}{bold}{\encodingdefault}{\sfdefault}{bx}{n}

\usepackage{hyperref}
\usepackage{url}
\usepackage[utf8]{inputenc} 
\usepackage[T1]{fontenc}    
\usepackage{booktabs}       
\usepackage{amsfonts}       
\usepackage{nicefrac}       
\usepackage{microtype}      
\usepackage{xcolor}         
\usepackage{colortbl}       
\usepackage{graphicx}
\usepackage{subcaption} 
\usepackage{enumitem}
\usepackage{tabularx}
\usepackage{multicol}
\usepackage{booktabs}
\usepackage{amssymb}
\usepackage{amsmath}
\usepackage{multirow}
\usepackage{placeins}
\usepackage{wrapfig}
\usepackage{needspace}
\usepackage{titletoc}
\definecolor{robotwinheader}{HTML}{C4DCEE}
\definecolor{liberoheader}{HTML}{DCE8B8}
\definecolor{robotwinheader}{HTML}{C4DCEE}
\definecolor{robotwinmixedheader}{HTML}{99BEDD}
\definecolor{robotwinzeroheader}{HTML}{E7F0F8}
\definecolor{liberoheader}{HTML}{DCE8B8}
\definecolor{oursrow}{HTML}{FFF3E6}
\definecolor{taskgroup}{HTML}{D8D8D8}
\definecolor{vlagroup}{HTML}{D9CDED}
\definecolor{wamgroup}{HTML}{C6E3DF}
\definecolor{wsagroup}{HTML}{F3D38A}
\newcommand{\eat}[1]{}

\newcommand{\resulttbd}[1]{}

\definecolor{CiteGreen}{HTML}{2E6B4F}
\hypersetup{
    colorlinks=true,
    citecolor=CiteGreen
}

\title{WSA$_1$: a 3D-Centric World-Spatial-Action Model for Generalizable Robot Control}

\author{%
	\textbf{Jiahao Jiang}   \quad 
	\textbf{Jianing Zhang}     \quad
	\textbf{Xuanhan Wang}\textsuperscript{$\diamond$}\quad
	\textbf{Duo Peng}\quad
	\textbf{Ruidong Chen}     \quad 
	\textbf{Zhenhan Yin}   \quad 
	\\
	[0.35em]
	\textbf{Sen Wang}   \quad 
	\textbf{Zhaoshu Yu}       \quad 
	\textbf{Pengpeng Zeng} \quad
	\textbf{Xiaofeng Cao} \quad 
	\textbf{Jingkuan Song}\textsuperscript{\dag}  
	\textbf{Heng Tao Shen}\textsuperscript{\dag}
	\\[0.35em]
	{Tongji University;} \quad
	{Shanghai Innovation Institution;} \quad
	{Shanghai Magic;} \quad
	{Koala Uran} 
	\\[0.35em]
	Project Page: \url{https://github.com/zaleni/WSA}
	\\[0.35em]
	\textsuperscript{\dag}Project Leader \quad
	\textsuperscript{$\diamond$}Corresponding Author
}
\iclrfinalcopy 
\begin{document}

\maketitle

\begin{abstract}
World-Action Models (WAMs) have emerged as a promising paradigm for robot control by using predicted world evolution to guide action generation. 
However, existing WAMs largely follow a one-way world-to-action paradigm. Real-world interaction, in contrast, is inherently driven by bidirectional temporal causality with implicit physics: the evolving world informs what actions should be taken, while the executed actions, in turn, determine how the world evolves. Without explicitly modeling this action-to-world direction, WAMs cannot assess whether the consequences of their generated actions align with the world evolution they predict, preventing WAMs from capturing consistent interaction dynamics from video demonstrations
and thereby constraining their generalization capabilities in real-world scenarios. 
In this paper, we introduce \textbf{WSA$_1$}, a new modeling paradigm for generalizable robot control. WSA$_1$ develops a World-Spatial-Action understanding by jointly modeling three interconnected processes: how actions transform the 3D world (world level), how spatial states evolve toward desired outcomes (spatial level), and how target 3D world guide appropriate robot behaviors (action level). By integrating these processes, WSA$_1$ enables more coherent and tightly coupled reasoning over embodied interactions.
Compared with prior WAMs trained on tens of thousands of hours of real-robot data, WSA$_1$ leverages 6k hours of video demonstrations (only 1k from real robot), demonstrating strong data efficiency. Moreover, even our 3B-parameter WSA$_1$ surpasses larger WAMs with 6B–8B parameters, highlighting its parameter efficiency. Experiments show that WSA$_1$ achieves a 93.3\% success rate on RoboTwin2.0, delivering an average 20\% improvement over state-of-the-art models on real-world robot control tasks.
\end{abstract}

\section{Introduction}

The pursuit of generalist robot policies has driven rapid progress in embodied foundation models, which seek to equip robots with transferable capabilities for interacting with diverse physical environments.
Among recent approaches, World-Action Models (WAMs)~\citep{wam:motus,wam:mimicvideo,wam:dreamzero,wam:cosmospolicy} have emerged as a promising paradigm for robot control.
Rather than generating actions solely from the current observation and task instruction, WAMs explicitly predict how the world may evolve and use the predicted future to guide action generation.
Such predictive modeling provides robots with an intermediate notion of ``what should happen next'' before deciding ``what to do next'', offering a potentially powerful basis for generalization beyond the training distribution.

However, existing WAMs largely formulate this relationship in only one direction: \emph{world-to-action}.
Predicted world evolution provides guidance for generating robot actions, yet the reverse direction---how those actions transform the world---is not explicitly modeled within the same interaction process.
This creates an important disconnect.
In real-world manipulation, world evolution and robot action are intrinsically coupled: the evolving world constrains which actions are appropriate, while the executed actions determine the subsequent world state.
A model that captures only the former can generate actions according to a predicted future, but lacks an explicit mechanism for learning whether those actions would actually induce a world transition consistent with that future.
As a result, world prediction and action generation are learned as only partially coupled components rather than two complementary views of the same physical interaction.

This observation motivates a different perspective on World-Action modeling: effective embodied learning should capture not only how a future world guides actions, but also how actions give rise to that future world.
Importantly, connecting these two directions requires more than independently learning a forward world model and an inverse action model.
Robot interaction is ultimately realized through changes in 3D spatial configurations: objects move, contacts change, and spatial relations evolve as actions are executed.
Such spatial evolution provides a common representation that connects an action with the world transformation it produces.
We therefore argue that \emph{world evolution, spatial evolution, and action generation should be modeled jointly}, rather than treating world prediction merely as an auxiliary conditioning signal for control.

\begin{figure*}[t]
	\centering
	\includegraphics[width=0.98\textwidth]{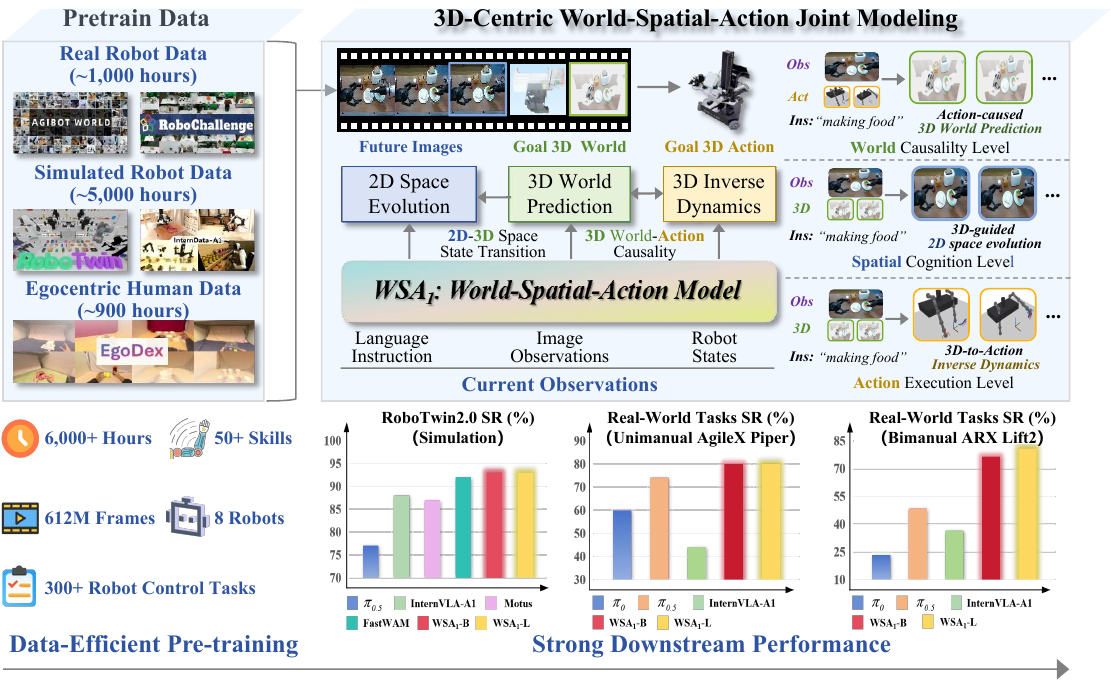}
	\captionof{figure}{
    \textbf{WSA$_1$}: A generalizable robot foundation model based on 3D-centric world-spatial-action joint modeling, which learns bidirectional world-action physical commonsense from real-world interactions. It achieves strong performance across diverse simulated and real-robot benchmarks using 6K hours of pre-training data (only 1k from real robot), demonstrating strong data efficiency.
    \vspace{-2mm}
	} 
	\label{fig:intro}
\end{figure*}

Based on this principle, we introduce \textbf{WSA$_1$}, a World-Spatial-Action modeling paradigm for generalizable robot control.
WSA$_1$ jointly captures three interconnected processes.
At the \emph{world level}, it models how robot actions transform the 3D world, learning the target world state induced by an interaction.
At the \emph{spatial level}, it models how spatial states evolve from the current observation toward the target 3D world, providing a geometric connection between action-induced changes and their resulting outcomes.
At the \emph{action level}, it models how the target 3D world guides appropriate robot behaviors, learning the inverse mapping from desired world evolution to executable actions.
Together, these processes connect action-to-world and world-to-action modeling through a shared spatial representation, allowing WSA$_1$ to learn more coherent world-action interaction dynamics from video demonstrations.

We instantiate WSA$_1$ as a unified embodied foundation model based on a Mixture-of-Transformers architecture, where complementary modeling experts jointly learn world evolution, spatial transformation, and action generation.
Rather than relying heavily on large-scale real-robot supervision, WSA$_1$ is trained on 6,000 hours of heterogeneous video demonstrations, of which only 1,000 hours are collected from real robots.
WSA$_1$ achieves a 93.3\% success rate on RoboTwin2.0 and improves over state-of-the-art approaches by an average of 20\% on real-world robot control tasks.
Moreover, our 3B-parameter model surpasses existing WAMs with 6B--8B parameters, demonstrating strong data and parameter efficiency.
In summary, our main contributions are three-fold:

\begin{itemize}[leftmargin=1.em]

\item \textbf{World-Spatial-Action Modeling.}
We introduce WSA$_1$, a 3D-centric world-action bidirectional learning paradigm that jointly models action-to-world evolution, spatial transformation, and world-to-action generation.

\item \textbf{Unified Embodied Foundation Model.}
    We instantiate WSA$_1$ with a Mixture-of-Transformers architecture that jointly optimizes world, spatial, and action modeling across two model scales.

\item \textbf{Efficient Generalization.}
    With 6,000 hours of demonstrations, including only 1,000 hours of real-robot data, WSA$_1$ achieves 93.3\% success on RoboTwin2.0, improves real-world performance by 20\% on average, and enables the 3B model to outperform larger 6B--8B WAMs.

\end{itemize}

\section{Related Work}

Embodied foundation models aim to transfer the broad visual and semantic knowledge of multi-modal foundation models~\citep{vlm:internvl,vlm:qwen25,video:wan} to robot perception and control~\citep{survey:zhang2025step,survey:yu2025survey}.
Existing approaches can be broadly grouped into Vision-Language-Action (VLA) models and World-Action Models (WAMs).

\noindent\textbf{Vision-Language-Action Models.}
VLA models~\citep{vla:dexvla,vla:pi0,vla:pi05,vla:egovla,vla:cotvla} leverage pretrained vision-language models and adapt their semantic priors to robot control through demonstration learning.
Early methods such as RT-2~\citep{vla:RT2} and OpenVLA~\citep{vla:openvla} formulate action generation as autoregressive token prediction, while recent approaches~\citep{vla:pi0,vla:pi05,vla:grootN} adopt continuous generative objectives such as flow matching for more precise control.
Several works further incorporate explicit 3D information, including depth and 3D representations, to enhance spatial reasoning~\citep{vla:3DVLA,vla:spatialvla}.
Despite these advances, VLA models primarily focus on learning observation-to-action mappings, whereas explicitly modeling how robot actions transform the future 3D world remains less explored.

\noindent\textbf{World-Action Models.}
WAMs~\citep{wam:mimicvideo,wam:dreamzero,wam:cosmospolicy,wam:fastwam,wam:mwm,wam:motus} incorporate predictive world modeling into robot control by jointly learning future world evolution and action generation, often leveraging pretrained video models as world priors.
Motus~\citep{wam:motus} demonstrates the benefit of future prediction for manipulation generalization, while Fast-WAM~\citep{wam:fastwam} shows that such gains can also arise from joint world-action learning without explicit video generation.
More recent approaches extend world prediction to 3D geometry~\citep{vla:geopredict,wam:dreamvla,wam:mvvdp,wam:pointworld}, enabling robot policies to exploit richer spatial information during manipulation.
However, existing WAMs predominantly model how predicted world evolution guides action generation, while the reverse direction---how robot actions induce corresponding changes in the 3D world---remains less explicitly modeled.

Compared with prior VLA and WAM approaches, our WSA$_1$ goes beyond direct observation-to-action mapping and one-way world-to-action prediction by additionally modeling how robot actions transform the 3D world.
It further connects these two directions through spatial-state evolution, jointly modeling action-to-world, spatial, and world-to-action dynamics within a unified framework.

\section{Methodology}
\subsection{Problem Definition and Notation}
In the context of robot control, the goal of embodied foundation model is to learn a robust policy that predicts next robot actions from current observations. Formally, at each timestep $t$, the observations include task instruction termed $l$, visual observation denoted as $v_t$, and robot proprioception termed $s_t$. In general, the robot action is a $H$-step chunk termed $A_{t}:=
({a}_{t+1},\ldots,{a}_{t+H})$, a short-term action trajectory with a sequence of end-effector pose or joints of a robot in next $H$ timesteps. To train a robot policy $\pi_{\theta}$ parameterized by $\theta$ on diverse video demonstrations $\mathcal{D}_{\text{dem}}$, it is required to maximize the standard likelihood objective:
\begin{equation}
	\max_{\theta} \mathbb{E}_{(v_t,s_t,A_{t},l) \sim \mathcal{D}_{\text{dem}}} \log \pi_{\theta}(A_{t} \mid v_t,s_t,l).
\end{equation}
Previous works formulate robot control problem as vision-language-action or world-action policy, which can be regarded as a joint distribution derived from the standard objective.
Different from previous VLA and WAM paradigms that rely on unidirectional mappings, we formulate robot control as world-spatial-action joint learning that captures the interdependencies among 3D world dynamics, visual space cognition, and robot actions. 
Specifically, it jointly learns three complementary conditional distributions that impose mutual constraints among 3D world model, 2D space transition, and 3D inverse dynamics model:
\begin{equation}
	\underbrace{p(g_{t+1:t+K} \mid a_{t+1:t+H}, \mathbf{O}_{t})}_{\text{Action-caused 3D World Model}}, \quad \underbrace{p(v_{t+1:t+N} \mid g_{t+1:t+K}, \mathbf{O}_{t})}_{\text{3D-guided 2D Space Transition}}, \quad 
	\underbrace{p(a_{t+1:t+H} \mid g_{t+1:t+K}, \mathbf{O}_{t})}_{\text{3D Inverse Dynamics Model}}
	\label{eq:3dwsa_policy}
\end{equation}
where $\mathbf{O}_{t}:=\{v_t,s_t,l\}$ denotes robot current observation, $\{{g}_{t+1},\ldots,{g}_{t+K}\}$ are target 3D world representations in near future, and $\{{v}_{t+1},\ldots,{v}_{t+N}\}$ represent visual states of images in near future. 
These three distributions correspond to the three levels of the WSA framework: the world causality level (predicting how actions shape the 3D world), the spatial cognition level (grounding visual predictions in 3D geometry), and the action execution level (generating motor commands from 3D world states).
\begin{figure*}[t]
	\centering
	\includegraphics[width=0.95\textwidth]{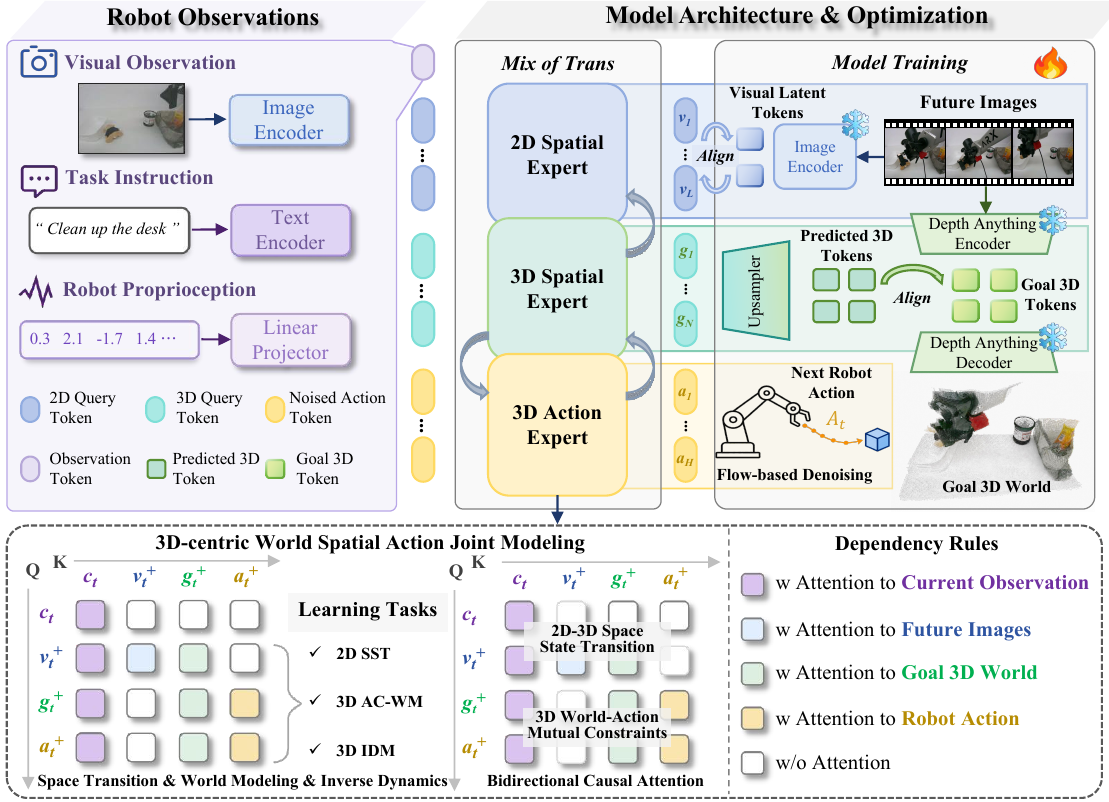}
	\captionof{figure}{Overview of WSA$_1$. It adopts a Mixture-of-Transformers structure with three complementary experts: a 2D Spatial Expert for 2D-3D space state transition, a 3D Spatial Expert for action-caused 3D world prediction, and a 3D Action Expert for 3D inverse dynamics modeling. A bidirectional causal attention mechanism enforces dependency rules to integrate world–action mutual constraints within a shared latent space.
	} 
	\label{fig:overview}
\end{figure*}
\subsection{Model Architecture} 
Inspired by the recent works~\citep{wam:motus,vla:pi05},  WSA$_1$ is built upon a mixture-of-transformer (MoT) architecture to enable seamless integration of 2D-3D space transition, 3D world prediction and robot action generation, involving 2D spatial expert in parallel with 3D spatial expert and 3D action expert.   

\textbf{2D Spatial Expert (2D-SE).} Given a human instruction, generating 3D goal-directed state transition in 2D space enhances understanding of world dynamics. The 2D-SE $f_{2D}(\cdot)$ serves as the 2D-3D state transition branch of WSA$_1$. It is initialized from a pre-trained vision-language model, thereby inheriting strong vision-language priors. 
The goal of 2D-SE is to predict future 2D visual dynamics to reach the target state of the 3D world. Specifically, given goal states $G_{t}:=({g}_{t+1},\ldots,{g}_{t+K})$ of 3D world, it predicts visual tokens $h_{v}=f_{2D}(G_{t}, O_{t})$ of those images in near future, instead of RGB pixels.
This design allows the model to focus on the essential visual dynamics rather than temporal variations of redundant pixels. 

\textbf{3D Spatial Expert (3D-SE).} Given a sequence of executable robot actions, predicting target world in 3D space reflects causal effect of the robot behavior. Under this context, the 3D-SE $f_{3D}(\cdot)$ is designed as an internal mechanism for 3D causal world modeling. Instead of directly construct native 3D representations such as 3D Gaussian or 3D point cloud, the 3D-SE predicts latent representations $h_{g}=f_{3D}(O_{t},A_{t})$ of target 3D world, which are derived from 3D foundation models such as DepthAnything~\citep{vfm:da3}. We instantiate 3D-SE with a lightweight transformer model of the same depth as 2D-SE, thereby the architecture of 3D-SE can be tightly coupled with 2D-SE for space state transition.

\textbf{3D Action Expert (3D-AE).} Given target 3D world, predicting next robot action tells why and how robot to behave in 3D space. The 3D-AE $f_{act}(\cdot)$ aims to predict a full action chunk conditioned on both current observations and future 3D latent representations. Specifically, we instantiate 3D-AE as denoising diffusion transformer. Given noisy actions $\hat{A}_{t} = (1-\tau)A_{t}+\tau\omega$, where $\omega \sim \mathcal{N}(0,I)$ and $\tau \in [0,1]$, it employs iterative cross-attention and denoising procedure to predict latent action representations $h_{act}=f_{act}(O_{t},G_{t},\hat{A}_{t})$.
This yields a 3D inverse dynamics formulation, where 3D-AE maps predicted 3D latent world representations back to executable action. 

\textbf{World-Spatial-Action Causal Attention.} Three experts are coupled within the mixture-of-transformer architecture via a shared attention mechanism. 
Formally, let ${h}_t$ denote tokens of current observation, which are extracted via 2D-SE.
As illustrated in Figure~\ref{fig:overview}, we adopt a 3D-centric causal attention mechanism, which defines the following dependency rules:
(1) All predicted tokens ($h_{v}$, $h_{g}$, $h_{act}$) can attend to the current observation tokens ${h}_t$, ensuring all predictions are grounded in the current state of the robot and the environment. 
(2) The visual tokens of future subgoal images $h_{v}$ can attend to the 3D scene tokens ${h}_{g}$, encouraging the model to learn 2D-3D visual consistency and ground semantic visual predictions in physically consistent 3D geometry.
(3) The 3D scene tokens $h_{g}$ and action tokens $h_{act}$ have full bidirectional attention access to each other, enabling joint learning of action-caused 3D world prediction and 3D goal-directed action generation. This bidirectional attention is the key to modeling world-action interdependence of the WSA paradigm, as it reflects a causal inductive bias that drives the model to capture how actions shape world evolution and how world states inform action generation, rather than merely learning correlational mappings. 

Based on these four components, we implement two variants of WSA$_1$, where their 2D-SE module is equipped with two different pretrained models and 3D-SE/AE is randomly initialized: WSA$_1$-B, a base model initialized from the pretrained Qwen3-VL-2B~\citep{vlm:qwen3} with a total of 3B parameters; and WSA$_1$-L, a large variant built on the pretrained Wan2.2-5B~\citep{video:wan} with a total of 6B parameters.

\subsection{World-Spatial-Action Joint Learning}

Analogous to the internal model theory in cognitive neuroscience~\citep{wolpert2000computational,clark2013whatever}, the WSA paradigm is designed to mirror three core computational principles of the human brain: 1) 3D spatial grounding of perception, 2) forward predictive modeling of action consequences, and 3) inverse motor control for goal-directed behavior. These three principles correspond to following three levels of the WSA learning framework, respectively.

\textbf{2D Space State Transition (2D SST).} Analogous to the brain's ability to form mental imagery of future scenes~\citep{kosslyn1996image}, the 2D Spatial Expert learns to predict 3D-grounded visual state transition. 
To this end, we adopt a MSE loss over the predicted visual tokens of future images, to minimize the discrepancy between the predicted image tokens and the ground-truth tokens extracted via pre-trained tokenizer. Formally, this transition learning objective is defined as:
\begin{equation}
	\mathcal{L}_{\text{2D}} = \mathbb{E}_{({O}_{t}, {G}_{t}, {V}_t) \sim \mathcal{D}_{dem}} \left\| f_{2D}(O_{t},G_{t}) - f_{enc}({V}_{t}) \right\|_2^2,
	\label{eq:loss_vgm}
\end{equation}
where $f_{enc}(\cdot)$ denotes a pre-trained VAE-based tokenizer (e.g., COSMOS~\citep{vfm:cosmos}).

\textbf{Action-Caused 3D World Modeling (3D AC-WM).} 
We adopt MSE loss over the predicted 3D world latent representations, to minimize the error between the predicted future 3D states and the ground-truth 3D geometric states on the demonstration data. This 3D world model learning objective can be formalized as:
\begin{equation}
	\mathcal{L}_{\text{3D}} = \mathbb{E}_{(O_t, A_t, G_t) \sim \mathcal{D}_{dem}} \left\| f_{3D}(O_{t},A_{t}) - f_g(G_t) \right\|_2^2,
	\label{eq:loss_3dvgm}
\end{equation}
where $f_g(\cdot)$ denotes a pre-trained 3D foundation model, e.g., Depth-Anything~\citep{vfm:da3}. This corresponds to the forward internal model in motor control theory~\citep{wolpert2000computational}, which predicts how motor actions will alter the state of the environment.

\textbf{3D Inverse Dynamics Modeling (3D IDM).} Following previous works~\citep{vla:geopredict,vla:pi05}, we adopt flow matching loss~\citep{diffusion:flow_matching} for continuous action generation. This loss function aims to minimize the error between the predicted flow field and the ground-truth one, conditioned on the current observation and the goal 3D world. This learning process can be formulated as:
\begin{equation}
	\mathcal{L}_{\text{ACT}} = \mathbb{E}_{(O_t,  G_t, A_t) \sim \mathcal{D}_{dem}} 
	\left\| \omega - A_{t} - f_{act}(O_{t},G_{t},\hat{A}_{t}) \right\|_2^2.
	\label{eq:loss_3dact}
\end{equation}
This trains 3D-AE to predict flow field ($\omega - A_{t}$), thereby facilitating coherent action generation.
The overall learning objective is the sum of the three losses:
\begin{equation}
	\mathcal{L}_{\text{total}} = \mathcal{L}_{\text{2D}} + \mathcal{L}_{\text{3D}} +  \mathcal{L}_{\text{ACT}}.
	\label{eq:total_loss}
\end{equation}

\begin{table}[h]
	\centering
	\small
	\setlength{\tabcolsep}{3.2pt}
	\renewcommand{\arraystretch}{1.15}
	\caption{Data mixture and training objectives used for pretraining.
		A checkmark indicates that the corresponding loss is enabled,
		while ``--'' indicates that the loss is masked out.}
	\label{tab:pretraining_data_mixture}
	\resizebox{0.75\linewidth}{!}{%
		\begin{tabular}{lcccccc}
			\toprule
			\textbf{Data Type}
			& \textbf{Num. frames}
			& \textbf{Num. tasks}
			& \textbf{Sampling weight}
			& ${\mathcal{L}_{2\mathrm{D}}}$
			& ${\mathcal{L}_{3\mathrm{D}}}$
			& ${\mathcal{L}_{\mathrm{ACT}}}$ \\
			\midrule
			Simulation Data
			& 424M & 70  & 0.54
			& $\checkmark$ & $\checkmark$ & $\checkmark$ \\
			Egocentric Human Data
			& 89M  & 194 & 0.10
			& $\checkmark$ & $\checkmark$ & -- \\			
			Real Robot Data
			& 99M & 36 & 0.36
			& $\checkmark$ & $\checkmark$ & $\checkmark$ \\
			\bottomrule
		\end{tabular}%
	}
\end{table}
\subsection{Pre-train Data}
\label{sec:pretrain_data}
High-quality embodied data is the cornerstone of pre-training generalizable embodied foundation models that can acquire versatile manipulation skills for flexible real-world robot control. Existing robot foundation models~\citep{vla:pi0,wam:lingbot-va2026} typically rely on tens of thousands of hours of robot demonstration data to achieve broad generalization, with real-robot robot data constituting a major portion of their pre-training corpora. 
However, as demonstrated in recent works~\citep{vla:grootN,vla:mivla}, data diversity is the key to robot foundation model pre-training. Accordingly, we carefully collect a pre-train dataset with only 6,000 hours of demonstration videos from a mixture of heterogeneous data sources, including simulation data~\citep{simulation_data:interns}, egocentric human data~\citep{human_data:egodex} and real robot data~\citep{data:agibot,data:robochallenge}. As illustrated in Table\ref{tab:pretraining_data_mixture}, the mixture of data sources covers 8 different robot embodiments and 300+ robot control tasks, including 50+ atomic manipulation skills such as grasping and placing. Since there exists embodiment gap between human and robot, we do not apply action generation loss ${\mathcal{L}_{\mathrm{ACT}}}$ to WSA models when pre-training on human data.

\section{Experiments}

We evaluate WSA$_1$ in both simulation and real-world settings, focusing on generalization,
the contribution of World-Spatial-Action joint modeling, and data efficiency.
We first investigate the merits of WSA modeling in simulation, and then examine whether the merits observed in simulation are consistent to that of the real robot setting. Together, these experiments provide a comprehensive validation of whether 3D-centric bidirectional world-spatial-action modeling offers a practical and data-efficient pathway to generalizable robot control. More results and analysis are in Appendix.

\subsection{Simulation Experiments}

\textbf{Benchmarks.}
We evaluate on RoboTwin2.0~\citep{simulation_data:robotwin2}, which contains 50 bimanual manipulation tasks under clean and randomized settings, and LIBERO~\citep{simulation_data:libero}, covering spatial, object, goal-conditioned, and long-horizon manipulation.
The randomized setting introduces scene-level distribution shifts and is used to evaluate out-of-distribution (OOD) generalization.

\textbf{Evaluation Protocol.}
We adopt multi-task evaluation with success rate (SR) as the metric.
Unless otherwise stated, models are fine-tuned on all 27,500 RoboTwin2.0 demonstrations.
For scene generalization (OOD-C2R), models are trained only on clean demonstrations and tested in randomized scenes.
For zero-shot cross-embodiment generalization (OOD-CE), the trained AgileX embodiment is replaced by an unseen ARX-X5 at test time.
By default, WSA$_1$-B without pre-training is adopted to conduct ablation studies for fast investigation.

\begin{table}[t]
	\centering
	\captionsetup{skip=4pt}
	\caption{Comparison on RoboTwin2.0 and LIBERO benchmarks.
		We report the mean success rate (SR) and the best results are shown in \textbf{bold}.}
	\scriptsize
	\label{tab:main_results}
	\setlength{\tabcolsep}{2.1pt}
	\renewcommand{\arraystretch}{1.05}
	\begin{tabularx}{\linewidth}{@{}
			>{\raggedright\arraybackslash}p{0.315\linewidth}
			>{\centering\arraybackslash}p{0.055\linewidth}
			*{7}{>{\centering\arraybackslash}X}@{}}
		\toprule
		\multirow{2}{*}[-1.4ex]{\textbf{Method}}
		& \multirow{2}{*}[-1.4ex]{\textbf{Size}}
		& \multicolumn{3}{c}{\cellcolor{robotwinheader}\textbf{RoboTwin2.0}}
		& \multicolumn{4}{c}{\cellcolor{liberoheader}\textbf{LIBERO}} \\
		\cmidrule(lr){3-5}\cmidrule(lr){6-9}
		&
		& \textbf{Clean}
		& \textbf{Rand.}
		& \textbf{Avg.}
		& \textbf{Spatial}
		& \textbf{Object}
		& \textbf{Goal}
		& \textbf{Long}
		\\
		\midrule
		\cellcolor{vlagroup}\textit{Vision--Language--Action (VLA) models}
		& \multicolumn{8}{c}{} \\
		OpenVLA~\citep{vla:openvla}
		& 7B & -- & -- & -- & 84.7 & 88.4 & 79.2 & 53.7   \\
		OpenVLA-OFT~\citep{vla:openvla_oft}
		& 7B & -- & -- & -- & 97.6 & 98.4 & 97.9 & 94.5   \\
		$\pi_0$~\citep{vla:pi0}
		& 3B & 65.9 & 58.4 & 62.2 & 96.8 & 98.8 & 95.8 & 85.2   \\
		$\pi_{0.5}$~\citep{vla:pi05}
		& 3B & 82.7 & 76.8 & 79.8 & 98.8 & 98.2 & 98.0 & 92.4   \\
		X-VLA~\citep{vla:xvla}
		& 1B & 72.8 & 72.8 & 72.8 & 98.2 & 98.6 & 97.8 & 97.6  \\
		InternVLA-A1~\citep{wam:internvla_a1}
		& 3B & 89.4 & 89.6 & 89.5 & -- & -- & -- & --   \\
		GR00T N1~\citep{vla:grootN}
		& 2B & -- & -- & -- & 95.6 & 97.6 & 94.2 & 89.6  \\
		SpatialVLA~\citep{vla:spatialvla}
		& 4B & -- & -- & -- & 88.2 & 89.9 & 78.6 & 55.5   \\
		HoloBrain-0-QW~\citep{vla:holobrain0}
		& 1B & 91.9 & 92.3 & 92.1 & 97.2 & 99.6 & 97.6 & 95.2  \\
		Pose-VLA (RGB)~\citep{vla:posevla}
		& 3B & 79.9 & 79.1 & 79.5 & 96.5 & 98.0 & 97.1 & 92.4   \\
		ABot-M0~\citep{vla:abot-m0}
		& 4B & 86.1 & 85.1 & 85.6 & 98.8 & 99.8 & \textbf{99.0} & 96.6  \\
		GeoPredict~\citep{vla:geopredict}
		& -- & -- & -- & -- & 98.0 & 98.2 & 95.7 & 94.0   \\
		\midrule
		\cellcolor{wamgroup}\textit{World--Action (WAM) models}
		& \multicolumn{8}{c}{} \\
		Motus~\citep{wam:motus}
		& 8B & 88.7 & 87.0 & 87.8 & 96.8 & 99.8 & 96.6 & 97.6  \\
		4D-WAM~\citep{wam:4dwam}
		& 6B & 91.9 & 90.8 & 91.3 & 98.8 & 99.5 & 98.6 & 97.4  \\
		Fast-WAM~\citep{wam:fastwam}
		& 6B & 91.9 & 91.8 & 91.8 & 98.2 & \textbf{100.0} & 97.0 & 95.2  \\
		LingBot-VA~\citep{wam:lingbot-va2026}
		& 5B & 92.9 & 91.6 & 92.2 & 98.5 & 99.6 & 97.2 & \textbf{98.5}  \\
		\midrule
		\cellcolor{wsagroup}\textit{World--Spatial--Action (WSA) models}
		& \multicolumn{8}{c}{} \\
		\rowcolor{oursrow}
		\textbf{WSA$_1$-B (Ours)}
		& 3B & 92.2 & 92.7 & 92.4 & 98.6 & 99.6 & 97.2 & 94.2   \\
		\rowcolor{oursrow}
		\textbf{WSA$_1$-L (Ours)}
		& 6B & \textbf{93.5} & \textbf{93.1} & \textbf{93.3}& \textbf{99.4} & 99.8 & 98.0 & 95.6
		\\
		\bottomrule
	\end{tabularx}
    \vspace{-4mm}
\end{table}

\textbf{Main Results on Simulation Benchmarks.} 
As shown in Table~\ref{tab:main_results}, WSA$_1$-L attains 93.5\%, 93.1\% and 93.3\% scores on RoboTwin-2.0 in the clean mode, in the randomized mode and on average, which is the highest average among all entries and 1.1 points above the strongest baseline, LingBot-VA (92.2\%). 
The gap between the clean and the randomized mode is 0.4 points for WSA$_1$-L and 0.5 points for WSA$_1$-B, while it amounts to 7.5 points for $\pi_0$ and 5.9 points for $\pi_{0.5}$. WSA$_1$-B makes the comparison favourable on the parameter-efficiency side as well, since at 3B parameters it surpasses the 8B Motus (87.8\%), the 6B 4D-WAM (91.3\%) and the 6B Fast-WAM (91.8\%). 
On LIBERO, the comparison is less discriminative, because the strongest baselines already operate close to saturation. Averaged over the four suites, WSA$_1$-L reaches 98.2\% and WSA$_1$-B reaches 97.4\%, both are competitive with  8B Motus (97.7\%), 6B 4D-WAM (98.6\%), and the 6B Fast-WAM (97.6\%). WSA$_1$ therefore obtains the best accuracy--efficiency trade-off.

\textbf{Ablation Study of World-Action Bidirectional Modeling.}
As illustrated in Table~\ref{tab:direction_ablation}, bidirectional world-action modeling improves the average success rate from 73.5\% to 80.2\%.
Removing either direction degrades performance: removing world-to-action reduces the score to 71.0\%, while removing action-to-world yields 73.5\%.
Notably, the latter matches the fully unidirectional setting, suggesting that the gain comes from explicitly coupling action-conditioned world prediction with world-guided action generation, rather than simply combining two prediction targets. Under the causal ordering implemented by the unidirectional variant, the predicted world is not conditioned on the actions, which is why that variant behaves like the model whose action-to-world direction has been removed.

\textbf{Ablation Study of Learning Objectives.}
As reported in Table~\ref{tab:learning_ablation}, optimizing the action objective alone yields 59.7\% average SR.
Adding 2D SST or 3D AC-WM improves it to 75.8\% and 74.1\%, respectively, while jointly optimizing all three objectives further reaches 80.2\%.
These results show that the spatial and world objectives provide complementary supervision for action learning, with consistent gains also observed in the randomized setting.

\begin{table}[t]
	\centering
	\begin{minipage}[t]{0.49\linewidth}
		\vspace{0pt}
		
		\centering
		\captionsetup{font=small,skip=4pt,justification=raggedright,
			singlelinecheck=false}
		\caption{Ablation study of bidirectional world-action modeling on RoboTwin2.0.}
		\label{tab:direction_ablation}
		\fontsize{7.7pt}{9pt}\selectfont
		\setlength{\tabcolsep}{2.2pt}
		\renewcommand{\arraystretch}{1.08}
		\begin{tabular*}{\linewidth}{@{\extracolsep{\fill}}lcccc@{}}
			\toprule
			\textbf{Setting} & \textbf{Clean} & \textbf{Rand} & \textbf{Avg.} & \textbf{Drop.} \\
			\midrule
			unidirectional WSA & 72.5 & 74.5 & 73.5 & --\\
			\midrule
			bidirectional WSA & \textbf{80.7} & \textbf{79.6} & \textbf{80.2} & -- \\
			w/o action-to-world & 72.6 & 74.4 & 73.5 & $\downarrow$ 6.7  \\
			w/o world-to-action & 70.4 & 71.6  &  71.0 & $\downarrow$ 9.2  \\
			\bottomrule
		\end{tabular*}
		
	\end{minipage}\hfill
	\begin{minipage}[t]{0.49\linewidth}
		\vspace{0pt}
		
		\centering
		\captionsetup{font=small,skip=4pt,justification=raggedright,
			singlelinecheck=false}
		\caption{Ablation study of learning objectives on RoboTwin2.0.}
		\label{tab:learning_ablation}
		\fontsize{7.7pt}{9pt}\selectfont
		\setlength{\tabcolsep}{2.2pt}
		\renewcommand{\arraystretch}{1.08}
		\begin{tabular*}{\linewidth}{@{\extracolsep{\fill}}lcccrrr@{}}
			\toprule
			\textbf{Setting} & $\mathcal{L}_{\mathrm{ACT}}$ & $\mathcal{L}_{\mathrm{2D}}$ & $\mathcal{L}_{\mathrm{3D}}$ & \textbf{Clean} & \textbf{Rand} & \textbf{Avg.} \\
			\midrule
			Action only & $\checkmark$ & -- & -- & 59.5 & 59.9 & 59.7 \\
			\midrule
			w SST & $\checkmark$ & $\checkmark$ & -- & 75.8 & 75.8 & 75.8 \\
			w AC-WM & $\checkmark$ & -- & $\checkmark$ & 73.4 & 74.7 & 74.1 \\
			Full WSA & $\checkmark$ & $\checkmark$ & $\checkmark$ & \textbf{80.7} & \textbf{79.6} & \textbf{80.2} \\
			\bottomrule
		\end{tabular*}
	\end{minipage}
\end{table}

\textbf{Investigation of WSA Pre-train/Post-train.}
As shown in Table~\ref{tab:pretraining_ablation}, we compare three ways of applying the WSA objectives across pre-training and post-training.
Using WSA only during post-training yields poor OOD transfer, with 5.4\% under randomized OOD-C2R and 6.6\% under randomized OOD-CE, indicating that the three-level objectives alone are insufficient to handle substantial distribution shifts.
Introducing WSA during pre-training substantially improves generalization, 
\begin{wraptable}{r}{0.50\textwidth}
	\centering
	\captionsetup{font=small,skip=4pt,justification=raggedright,
		singlelinecheck=false}
	\caption{Investigation of WSA joint modeling for pre-training and post-training.}
	\label{tab:pretraining_ablation}
	\fontsize{7.8pt}{5pt}\selectfont
	\setlength{\tabcolsep}{2.35pt}
	\renewcommand{\arraystretch}{1.8}
	\begin{tabularx}{\linewidth}{@{}ll*{4}{>{\centering\arraybackslash}X}@{}}
		\toprule
		\multirow{2}{*}{\textbf{Pre-train}}
		& \multirow{2}{*}{\textbf{Post-train}}
		& \multicolumn{2}{c}{\cellcolor{robotwinheader}\textbf{OOD-C2R}}
		& \multicolumn{2}{c}{\cellcolor{robotwinzeroheader}\textbf{OOD-CE}} \\
		\cmidrule(lr){3-4}\cmidrule(lr){5-6}
		& 
		& \textbf{Clean} & \textbf{Rand}
		& \textbf{Clean} & \textbf{Rand} \\
		\midrule
		None & WSA  & 61.1 & 5.4 & 8.2 & 6.6 \\
		WSA & Action only & 83.4 & 75.9 & 46.7 & 46.4 \\
		WSA & WSA 
		& \textbf{83.8} & \textbf{77.8} & \textbf{47.1} & \textbf{47.6} \\
		\bottomrule
	\end{tabularx}
\end{wraptable}
raising randomized OOD-C2R performance to 75.9\% and clean performance from 61.1\% to 83.4\%.
Retaining WSA objectives during post-training further achieves the best overall results (83.8\%, 77.8\%, 47.1\%, and 47.6\%).
These results indicate that WSA-based pre-training learns transferable physical interaction priors, while preserving the objectives during post-training further strengthens their transfer to unseen scenes and embodiments.

\begin{table}[ht]
	\centering
	\begin{minipage}[t]{0.49\linewidth}
		\vspace{0pt}
		\centering
		\captionsetup{font=small,skip=4pt,justification=raggedright,
			singlelinecheck=false}
		\caption{Investigation of scene generalization. All models are fine-tuned with clean data only.}
		\label{tab:ood-c2r}
		\fontsize{7.7pt}{9pt}\selectfont
		\setlength{\tabcolsep}{2.2pt}
		\renewcommand{\arraystretch}{1.08}
		\begin{tabular*}{\linewidth}{@{\extracolsep{\fill}}lccc@{}}
			\toprule
			\textbf{Method} & \textbf{Clean} & \textbf{Rand} & \textbf{Avg.} \\
			\midrule
			$\pi_{0.5}$~\citep{vla:pi05} & 70.7 & 46.0 & 58.4 \\
			ABot-M0~\citep{vla:abot-m0} & 57.4 & 30.4 & 43.9 \\
			RobotManip~\citep{vla:qwen_robotmanip} & 73.2 & 62.6 & 67.9 \\
			Fast-WAM~\citep{wam:fastwam} & 77.8 & 1.90 & 39.9 \\
			InternVLA-A1~\citep{wam:internvla_a1} & \textbf{86.6} & 74.6 & 80.6 \\
			\midrule
			WSA$_1$ & 83.8 & \textbf{77.8} & \textbf{80.8} \\
			\bottomrule
		\end{tabular*}
	\end{minipage}\hfill
	\begin{minipage}[t]{0.49\linewidth}
		\vspace{0pt}
		\centering
		\captionsetup{font=small,skip=4pt,justification=raggedright,
			singlelinecheck=false}
		\caption{Investigation of zero-shot cross embodiment generalization. Unseen ARX-X5 is adopted for testing.}
		\label{tab:ood-ce}
		\fontsize{7.7pt}{9pt}\selectfont
		\setlength{\tabcolsep}{2.2pt}
		\renewcommand{\arraystretch}{1.08}
		\begin{tabular*}{\linewidth}{@{\extracolsep{\fill}}lccc@{}}
			\toprule
			\textbf{Method}  & \textbf{Clean} & \textbf{Rand} & \textbf{Avg.} \\
			\midrule
			$\pi_{0.5}$~\citep{vla:pi05} & 33.5 & 19.0 & 26.3 \\
			LingBot-VA~\citep{wam:lingbot-va2026} & 17.2 & 16.6 & 16.9 \\
			RobotManip~\citep{vla:qwen_robotmanip} & -- & 42.9 & 42.9 \\
			Fast-WAM~\citep{wam:fastwam} & 25.9 & 22.4 & 24.2 \\
			InternVLA-A1~\citep{wam:internvla_a1} & 18.1 & 17.5 & 17.8\\
			\midrule
			WSA$_1$ & \textbf{47.1} & \textbf{47.6} & \textbf{47.4} \\
			\bottomrule
		\end{tabular*}	
	\end{minipage}
\end{table}
\textbf{Scene Generalization.}
As shown in Table~\ref{tab:ood-c2r}, all models are fine-tuned on clean demonstrations and tested in randomized scenes to evaluate scene-level generalization.
WSA$_1$-B achieves the highest randomized success rate (77.8\%) and the highest average (80.8\%), outperforming InternVLA-A1 by 3.2 points in the randomized setting.
More importantly, WSA$_1$-B shows only a 6.0-point drop from clean to randomized scenes, substantially smaller than those of the competing methods.
For example, InternVLA-A1 drops by 12.0 points, while Fast-WAM decreases from 77.8\% to 1.9\%.
These results suggest that WSA$_1$ is more robust to scene-level distribution shifts and less dependent on appearance-specific correlations.

\textbf{Cross-embodiment Generalization.}
As shown in Table~\ref{tab:ood-ce}, all models are fine-tuned on the source embodiment and evaluated on an unseen ARX-X5 robot without target-platform fine-tuning.
WSA$_1$-B achieves 47.1\% and 47.6\% in the clean and randomized modes, respectively, averaging 47.4\%.
It outperforms $\pi_{0.5}$ by 13.6 and 28.6 points in the two settings and exceeds the best remaining average, RobotManip (42.9\%), by 4.5 points.
Nevertheless, the absolute performance remains well below the in-distribution result in Table~\ref{tab:main_results}, indicating that zero-shot cross-embodiment transfer remains challenging despite the clear relative gains.

\textbf{Summary of Exploration Study.}
Overall, the experiments provide two main findings:
1) \textit{Generalization.} WSA$_1$ consistently performs strongly across in-distribution, scene generalization (OOD-C2R), and zero-shot cross-embodiment (OOD-CE) settings, while remaining parameter-efficient;
2) \textit{Modeling effectiveness.} What makes 3D world modeling useful for robot control is the bidirectional joint modeling, not the objective itself.
This bidirectional world-action modeling is where generalization comes from, and it must be acquired during pre-training.

\begin{figure}[!t]
	\centering
	\includegraphics[width=\linewidth]{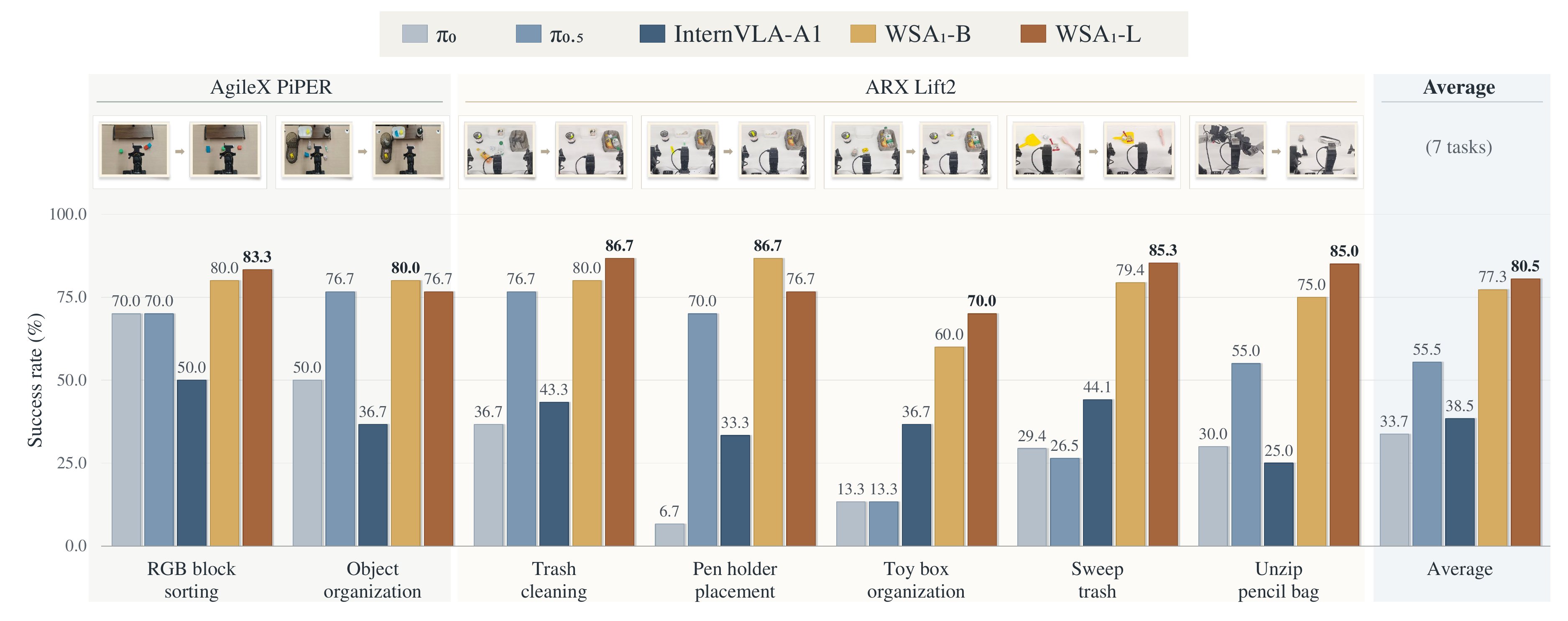}
    \vspace{-6mm}
	\caption{\textbf{Real-robot success rates across seven tasks.} Bars compare
		five policies on each task and their mean success rate over all seven tasks.
		Thumbnails show initial and final states. The first two tasks use AgileX
		PiPER, and the remaining five use ARX Lift2.
        }
	\label{fig:real_robot_bars}
\end{figure}

\subsection{Real Robot Experiments}

\textbf{Experimental Setup.}
Real-robot experiments are conducted on two platforms: (1) a unimanual AgileX PiPER with three cameras and a 7-DoF arm, and (2) a bimanual ARX Lift2 with three cameras and two 7-DoF arms.
We design seven tabletop manipulation tasks and evaluate all policies with objects initialized in predefined positions, orientations, and scene layouts to evaluate generalization capability.
We report two metrics: Success Rate (SR), the proportion of successfully completed rollouts; and Completeness (C), the proportion of completed sub-tasks.

\textbf{Baselines.}
We compare WSA$_1$ with three representative open-source embodied foundation models: $\pi_0$~\citep{vla:pi0} for continuous action prediction, $\pi_{0.5}$~\citep{vla:pi05} for open-world robot control, and InternVLA-A1~\citep{wam:internvla_a1}, which unifies visual understanding, video generation, and action generation.

\textbf{Main Results on Real-world Manipulation Tasks.}
As shown in Figure~\ref{fig:real_robot_bars}, WSA$_1$ achieves leading performance over baselines in the real-world setting.
Across 7 tabletop manipulation tasks, WSA$_1$-L achieves an average success rate of 80.5\% and an average completeness score of 85.2\%, while WSA$_1$-B reaches 77.3\% and 82.3\%, substantially outperforming the second-best model $\pi_{0.5}$ with 55.5\% and 63.3\%, respectively.

The per-task results further show that the advantage becomes more pronounced on challenging bimanual tasks.
On the two unimanual PiPER tasks, the baselines remain competitive, with $\pi_{0.5}$ reaching 70.0\% on \textit{RGB Block Sorting} and 76.7\% on \textit{Object Organization}.
On the five bimanual ARX Lift2 tasks, however, both WSA$_1$ models achieve at least 60.0\% success on every task, while $\pi_0$ and InternVLA-A1 remain below 50.0\% throughout.
The largest gaps appear on tasks requiring long-horizon execution and precise spatial reasoning: WSA$_1$-B reaches 86.7\% on \textit{Pen Holder Placement} versus 6.7\% for $\pi_0$, while WSA$_1$-L achieves 70.0\% on \textit{Toy Box Organization}, 85.0\% on \textit{Unzip Pencil Bag}, and 85.3\% on \textit{Sweep Trash}, substantially exceeding the baselines.

These results highlight the effectiveness of the proposed WSA architecture.
By coupling 3D-aligned visual state transition, 3D world dynamics, and 3D-conditioned action generation, WSA$_1$ enables more reliable execution in spatially complex and long-horizon tasks.

\section{Conclusion}

This work proposes WSA, a 3D-centric world-spatial-action paradigm that jointly models 3D space state transition, action-caused 3D world prediction, and 3D inverse dynamics within a Mixture-of-Transformers architecture.
We instantiate WSA at two scales, i.e., WSA$_1$-B (3B) and WSA$_1$-L (6B).
With only 6,000 hours of demonstrations (1,000 hours from real robots), WSA achieves strong performance across simulation and real-world benchmarks, demonstrating a data-efficient path toward generalizable robot foundation models.

\section*{AI Use Statement}
In the preparation of this paper, we employed generative AI solely as a writing
assistance tool for limited text polishing and language refinement. These AI tools were not involved in any
aspects of research ideation, conceptual development, technical analysis, algorithm design, experimental execution, or result analyses. All scientific contributions, methodological innovations, and intellectual content remain entirely our own.

\bibliography{iclr2027_conference}
\bibliographystyle{iclr2027_conference}

\clearpage
\appendix
\raggedbottom
\captionsetup{font=small}
\setlength{\textfloatsep}{10pt plus 2pt minus 2pt}
\setlength{\floatsep}{8pt plus 2pt minus 2pt}
\setlength{\intextsep}{8pt plus 2pt minus 2pt}
\renewcommand{\floatpagefraction}{0.75}
\renewcommand{\bottomfraction}{0.8}
\setcounter{topnumber}{3}
\setcounter{bottomnumber}{2}
\setcounter{totalnumber}{5}
This appendix records the configuration behind every experiment illustrated in the main paper and adds more evidence to support core claims in this paper. The first part specifies how the two WSA models were pre-trained and fine-tuned, how the world-modeling objectives were supervised, how large the deployed models are and what they cost at inference time, and the protocol under which every evaluation was run. The second part reports the full benchmark comparisons, the two distribution-shift studies, the directional ablation together with its qualitative evidence, and the real-robot task suite with its per-task scores.

\section{Implementation Details}
\label{app:experimental_details}
\Needspace{5\baselineskip}
\subsection{Pre-training configuration}

WSA$_1$-B and WSA$_1$-L share an optimisation schedule of 300,000 training steps at an effective batch size of 96 on eight GPUs, with cosine decay to a final learning rate of 1$\times$$10^{-5}$, bfloat16 precision and random seed 42. Three things differ between the scales, and Table~\ref{tab:pretraining_hparams} makes them explicit. The 2D Spatial Expert is initialized from a different backbone, Qwen3-VL-2B-Instruct for WSA$_1$-B against Wan2.2-TI2V-5B for WSA$_1$-L, which also changes the frozen image tokenizer each model carries. The visual and textual context is larger for WSA$_1$-L, which is pre-trained at 384$\times$320 with a 128-token text context while WSA$_1$-B uses 224$\times$224 and 48 tokens, so the larger model runs a higher peak learning rate (1$\times$$10^{-4}$ against 5$\times$$10^{-5}$) with a longer warmup (5,000 against 2,000 steps). The loss weighting is scale-specific as well: ($\lambda_{2D}$, $\lambda_{3D}$, $\lambda_{ACT}$) is (0.01, 0.01, 1.0) for WSA$_1$-B and (1.0, 0.1, 1.0) for WSA$_1$-L, so the 2D state-transition objective carries two orders of magnitude more weight at the larger scale, while the action objective stays at 1.0 at both. Actions are represented as deltas with a horizon of 50 for WSA$_1$-B and 32 for WSA$_1$-L, and inference uses 10 flow-matching steps.

\begin{table}[!htbp]
    \centering
    \small
    \setlength{\tabcolsep}{5pt}
    \renewcommand{\arraystretch}{1.10}
    \caption{Pretraining hyperparameters for WSA$_1$-B and WSA$_1$-L. Loss
    weights are ordered as $(\lambda_{2\mathrm{D}},\lambda_{3\mathrm{D}},
    \lambda_{\mathrm{ACT}})$.}
    \label{tab:pretraining_hparams}
    \begin{tabularx}{\linewidth}{@{}l>{\centering\arraybackslash}X>{\centering\arraybackslash}X@{}}
        \toprule
        \textbf{Configuration} & \textbf{WSA$_1$-B (3B)}
        & \textbf{WSA$_1$-L (6B)} \\
        \midrule
        Backbone initialization & Qwen3-VL-2B-Instruct & Wan2.2-TI2V-5B \\
        Optimizer & AdamW & AdamW \\
        AdamW $(\beta_1,\beta_2,\epsilon)$
        & $(0.9,0.95,10^{-8})$ & $(0.9,0.95,10^{-8})$ \\
        Weight decay & 0.01 & 0.01 \\
        Learning-rate schedule & Warmup + cosine decay & Warmup + cosine decay \\
        Peak $\rightarrow$ final learning rate
        & $5\!\times\!10^{-5}\rightarrow1\!\times\!10^{-5}$
        & $1\!\times\!10^{-4}\rightarrow1\!\times\!10^{-5}$ \\
        Warmup steps & 2,000 & 5,000 \\
        Training steps & 300,000 & 300,000 \\
        Per-device batch & 12 & 12 \\
        Number of GPUs & 8 & 8 \\
        Gradient accumulation & 1 & 1 \\
        Effective global batch & 96 & 96 \\
        Numerical precision & bfloat16 & bfloat16 \\
        Random seed & 42 & 42 \\
        Text context length & 48 & 128 \\
        Visual input resolution & $224\!\times\!224$ & $384\!\times\!320$ \\
        Action representation & Delta & Delta \\
        Action horizon & 50 & 32 \\
        Flow-matching inference steps & 10 & 10 \\
        Loss weights $(2\mathrm{D}/3\mathrm{D}/\mathrm{ACT})$
        & $0.01/0.01/1.0$ & $1.0/0.1/1.0$ \\
        \bottomrule
    \end{tabularx}
\end{table}

\Needspace{5\baselineskip}
\subsection{Supervision of the 3D world modeling objective}
The 3D world modeling objective is supervised by a frozen DA3-Large-1.1 teacher \citep{vfm:da3}. Each of the three camera views is processed at 504$\times$504, and target features are read from four hidden layers of the teacher, 11, 15, 19 and 23, with indices starting at zero. Four layers of the WSA$_1$ backbone are aligned to them through a single cross-attention block followed by a SiLU feed-forward network, layers 13, 19, 23 and 27 for WSA$_1$-B and 14, 19, 24 and 29 for WSA$_1$-L. Every view contributes 144 scene tokens and 1,296 target features arranged on a 36$\times$36 grid, all projected to a feature dimension of 2,048. Because the teacher is frozen, these objectives inherit only the structure that its representations contain, and no part of it is loaded at inference. Table~\ref{tab:3d_supervision_config} summarizes the 3D supervision configuration.

\begin{table}[!htbp]
    \centering
    \small
    \setlength{\tabcolsep}{4pt}
    \renewcommand{\arraystretch}{1.10}
    \caption{3D supervision configuration during pretraining. Layer indices
    start at zero.}
    \label{tab:3d_supervision_config}
    \begin{tabularx}{\linewidth}{@{}l>{\centering\arraybackslash}X>{\centering\arraybackslash}X@{}}
        \toprule
        \textbf{Configuration} & \textbf{WSA$_1$-B} & \textbf{WSA$_1$-L} \\
        \midrule
        DA3 specification & \multicolumn{2}{c@{}}{DA3-Large-1.1} \\
        input resolution per view & \multicolumn{2}{c@{}}{$504\times504$} \\
        selected hidden layers in DA3 & \multicolumn{2}{c@{}}{$11,15,19,23$} \\
        selected hidden layers in WSA$_1$ & $13,19,23,27$ & $14,19,24,29$ \\
        Feature dimension & \multicolumn{2}{c@{}}{2,048} \\
        Camera views & \multicolumn{2}{c@{}}{3} \\
        Scene tokens per view & \multicolumn{2}{c@{}}{144} \\
        Target features per view & \multicolumn{2}{c@{}}{1,296 ($36\times36$ grid)} \\
        Feature alignment module & \multicolumn{2}{c@{}}{One cross-attention block + SiLU FFN} \\
        \bottomrule
    \end{tabularx}
\end{table}

\Needspace{5\baselineskip}
\subsection{Downstream fine-tuning}
Fine-tuning uses one recipe across every benchmark and platform: AdamW with ($\beta_1$, $\beta_2$, $\epsilon$) = (0.9, 0.95, $10^{-8}$), weight decay 0.01, cosine learning-rate decay, bfloat16 precision, seed 42 and eight GPUs, with 10 flow-matching steps at inference. Table~\ref{tab:finetuning_hparams} lists what that recipe is instantiated with, namely the platform a policy is trained on, the schedule length, the effective batch size, the peak and final learning rates, the horizon H of predicted actions together with the number E of actions executed before replanning, the action representation and the loss weights. Three of these depend on the benchmark rather than on the model scale. The action representation is delta everywhere except LIBERO with WSA$_1$-L, which uses absolute actions. The execution horizon E is 24 on RoboTwin2.0, 10 on LIBERO and 50 or 48 on the real robots. And the 2D and 3D loss weights are re-tuned per benchmark, from (0.005, 0.01, 1.0) on LIBERO with WSA$_1$-B to (1.0, 0.1, 1.0) on RoboTwin2.0 with WSA$_1$-L, while $\lambda_{ACT}$ remains 1.0 in every row. 

\begin{table}[!htbp]
    \centering
    \footnotesize
    \setlength{\tabcolsep}{1.8pt}
    \renewcommand{\arraystretch}{1.08}

    \caption{
    Hyperparameters for downstream adaptation. Batch denotes the effective
    global batch size.
    Loss weights are ordered as
    $(\lambda_{2\mathrm{D}},\lambda_{3\mathrm{D}},\lambda_{\mathrm{ACT}})$.
    }
    \label{tab:finetuning_hparams}

    \begin{tabularx}{\linewidth}{@{}
        >{\raggedright\arraybackslash}m{0.105\linewidth}
        >{\centering\arraybackslash}m{0.090\linewidth}
        >{\centering\arraybackslash}m{0.110\linewidth}
        >{\centering\arraybackslash}m{0.055\linewidth}
        >{\centering\arraybackslash}m{0.100\linewidth}
        >{\centering\arraybackslash}m{0.100\linewidth}
        >{\centering\arraybackslash}m{0.075\linewidth}
        >{\centering\arraybackslash}X
        >{\centering\arraybackslash}m{0.145\linewidth}
        @{}}

        \toprule

        \textbf{Model}
        & \textbf{Platform}
        & \textbf{Duration}
        & \textbf{Batch}
        & \textbf{Peak}
        & \textbf{Final}
        & \textbf{Warmup}
        & \textbf{Action}
        & \textbf{Loss weights}
        \\[-1.5pt]

        &
        &
        &
        &
        \textbf{LR}
        & \textbf{LR}
        &
        & $(H/E)$
        &
        \\

        \midrule
        \multicolumn{9}{@{}l}{\textit{\textbf{RoboTwin2.0}}}
        \\[1pt]

        WSA$_1$-B
        & Aloha
        & 200k steps
        & 128
        & $5\times10^{-5}$
        & $5\times10^{-6}$
        & 2,000
        & Delta (50/24)
        & $0.01/0.01/1.0$
        \\

        WSA$_1$-L
        & Aloha
        & 5 epochs
        & 96
        & $1\times10^{-4}$
        & $1\times10^{-6}$
        & 5,000
        & Delta (32/24)
        & $1.0/0.1/1.0$
        \\

        \midrule
        \multicolumn{9}{@{}l}{\textit{\textbf{LIBERO}}}
        \\[1pt]

        WSA$_1$-B
        & Franka
        & 128k steps
        & 128
        & $7.5\times10^{-5}$
        & $5\times10^{-6}$
        & 2,000
        & Delta (10/10)
        & $0.005/0.01/1.0$
        \\

        WSA$_1$-L
        & Franka
        & 110k steps
        & 128
        & $7.5\times10^{-5}$
        & $5\times10^{-6}$
        & 2,000
        & Abs. (12/10)
        & $0.1/0.1/1.0$
        \\

        \midrule
        \multicolumn{9}{@{}l}{\textit{\textbf{Real robot}}}
        \\[1pt]

        WSA$_1$-B
        & PiPER
        & 30k steps
        & 128
        & $5\times10^{-5}$
        & $5\times10^{-6}$
        & 600
        & Delta (50/50)
        & $0.01/0.01/1.0$
        \\

        WSA$_1$-B
        & Lift2
        & 40k steps
        & 128
        & $5\times10^{-5}$
        & $5\times10^{-6}$
        & 600
        & Delta (50/50)
        & $0.01/0.01/1.0$
        \\

        WSA$_1$-L
        & PiPER
        & 40k steps
        & 96
        & $6\times10^{-5}$
        & $1\times10^{-5}$
        & 600
        & Delta (48/48)
        & $0.1/0.1/1.0$
        \\

        WSA$_1$-L
        & Lift2
        & 50k steps
        & 96
        & $6\times10^{-5}$
        & $1\times10^{-5}$
        & 600
        & Delta (48/48)
        & $0.1/0.1/1.0$
        \\

        \bottomrule
    \end{tabularx}
\end{table}

\Needspace{5\baselineskip}
\subsection{Parameter budget and inference cost}
Table~\ref{tab:optimization_scope} splits the trained parameters by expert. WSA$_1$-B holds 3.156B: 2.128B in the 2D Spatial Expert, 0.440B in the 3D Spatial Expert, 0.440B in the 3D Action Expert and about 0.148B in the remainder. WSA$_1$-L holds 6.457B, with 5.000B, 0.436B and 1.021B in the three experts. The growth from the smaller to the larger scale is therefore concentrated in the 2D Spatial Expert, which supplies 2.872B of the 3.301B added, or 87\% of the increase. Every one of these parameters is updated during training, and the frozen components, namely the image tokenizer and the frozen 3D teacher backbone, are listed separately and excluded from the totals.

\begin{table}[!htbp]
    \centering
    \footnotesize
    \setlength{\tabcolsep}{3pt}
    \renewcommand{\arraystretch}{1.15}
    \caption{The statistics of model parameters}
    \label{tab:optimization_scope}
    \begin{tabular*}{\linewidth}{@{\extracolsep{\fill}}lccccc@{}}
        \toprule
        \textbf{Model}
        & \textbf{2D-SE}
        & \textbf{3D-SE}
        & \textbf{3D-AE}
        & \textbf{Other}
        & \textbf{Total} \\
        \midrule
        WSA$_1$-B
        & 2.128B & 0.440B & 0.440B & $\sim$0.148B & 3.156B \\
        WSA$_1$-L
        & 5.000B & 0.436B & 1.021B & $\sim$0.1M & 6.457B \\
        \midrule
        \multicolumn{6}{@{}l}{\emph{Frozen components excluded from policy totals}} \\
        WSA$_1$-B
        & \multicolumn{5}{l@{}}{Cosmos image tokenizer (76.8M); DA3-Large backbone (304.4M)} \\
        WSA$_1$-L
        & \multicolumn{5}{l@{}}{Wan VAE (704.7M); DA3-Large backbone (304.4M)} \\
        \bottomrule
    \end{tabular*}
\end{table}

Table~\ref{tab:inference_footprint} measures a single action-inference call on one A800-SXM4-80GB with the frozen DA3 disabled. WSA$_1$-B peaks at 6.18 GiB and returns a chunk in 812.6 ms at the median and 870.1 ms at the 95th percentile, sustaining 1.210 chunks per second; WSA$_1$-L peaks at 14.01 GiB and returns a chunk in 1060.1 ms, sustaining 0.945 chunks per second. Doubling the trained parameter count therefore costs 2.3 times the peak memory and reduces chunk throughput by 22\%. The parameter count resident at inference is slightly above the trained total, 3.233B against 3.156B for WSA$_1$-B and 7.161B against 6.457B for WSA$_1$-L, and every row is measured with batch size 1 and 10 flow-matching steps.

\begin{table}[!htbp]
    \centering
    \footnotesize
    \setlength{\tabcolsep}{4pt}
    \renewcommand{\arraystretch}{1.12}
    \caption{Single-GPU action inference on an A800-SXM4-80GB with BF16, batch
    size 1, 10 denoising steps, the DA3 teacher disabled, 10 warmup runs, and
    50 measured runs.}
    \label{tab:inference_footprint}
    \begin{tabular*}{\linewidth}{@{\extracolsep{\fill}}lcccc@{}}
        \toprule
        \textbf{Model}
        & \shortstack{\textbf{Params (B)}\\\textbf{policy / inference}}
        & \shortstack{\textbf{Peak VRAM}\\\textbf{(GiB)}}
        & \shortstack{\textbf{Latency p50/p95}\\\textbf{(ms)}}
        & \textbf{Chunk/s} \\
        \midrule
        WSA$_1$-B & 3.156 / 3.233 & 6.18 & 812.6 / 870.1 & 1.210 \\
        WSA$_1$-L & 6.457 / 7.161 & 14.01 & 1060.1 / 1088.5 & 0.945 \\
        \bottomrule
    \end{tabular*}
\end{table}

\FloatBarrier
\section{Additional Results and Analyses}

\Needspace{5\baselineskip}
\subsection{Full simulation results}
\label{app:simulation_results}

Table~\ref{tab:libero_results} reproduces the complete LIBERO comparison behind the summary in Table 2, with 24 baselines drawn from both the VLA and the WAM families. The benchmark is close to saturation: 14 of the 26 entries average at least 97.0\% and six average at least 98.0\%, and the remaining headroom sits in the long-horizon suite, the only column in which the field is not compressed into the last two points. WSA$_1$-L averages 98.2\%, fifth among the 26 entries and 1.0 point below the best entry (Being-H0.7 at 99.2\%), while WSA$_1$-B averages 97.4\%; both sit inside the cluster of 6B and 8B models rather than above it. That is why Section 4.1 treats LIBERO as a consistency check and separates the two paradigms on RoboTwin2.0 and on the real robots.

\begin{table}[!htbp]
    \centering
    \caption{Full comparison with prior methods on LIBERO.  Dashes indicate unreported results.}
    \label{tab:libero_results}
    \footnotesize
    \setlength{\tabcolsep}{2.5pt}
    \renewcommand{\arraystretch}{1.10}
    \begin{tabular}{lccccc}
        \toprule
        \textbf{Method} & \textbf{Spatial} & \textbf{Object} & \textbf{Goal} & \textbf{Long} & \textbf{Avg.} \\
        \midrule
        ACT~\citep{robot-policy:act} & 96.3 & 99.5 & 97.0 & 90.8 & 95.9 \\
        Diffusion Policy~\citep{robot-policy:diffusion} & 78.3 & 92.5 & 68.3 & 50.5 & 72.4 \\
        Octo~\citep{vla:octo} & 78.9 & 85.7 & 84.6 & 51.1 & 75.1 \\
        OpenVLA~\citep{vla:openvla} & 84.7 & 88.4 & 79.2 & 53.7 & 76.5 \\
        $\pi_0$~\citep{vla:pi0} & 96.8 & 98.8 & 95.8 & 85.2 & 94.2 \\
        $\pi_{0.5}$~\citep{vla:pi05} & 98.8 & 98.2 & 98.0 & 92.4 & 96.9 \\
        OpenVLA-OFT~\citep{vla:openvla_oft} & 97.6 & 98.4 & 97.9 & 94.5 & 97.1 \\
        X-VLA~\citep{vla:xvla} & 98.2 & 98.6 & 97.8 & 97.6 & 98.1 \\
        GR00T N1~\citep{vla:grootN} & 95.6 & 97.6 & 94.2 & 89.6 & 94.3 \\
        GR00T N1.5~\citep{vla:grootN,vla:grootn15} & 98.2 & 99.4 & 97.2 & 87.8 & 95.7 \\
        GR00T N1.6~\citep{vla:grootN,vla:grootn16} & 97.7 & 98.5 & 97.5 & 94.4 & 97.0 \\
        SpatialVLA~\citep{vla:spatialvla} & 88.2 & 89.9 & 78.6 & 55.5 & 78.1 \\
        StarVLA-OFT (4B, spec.)~\citep{vla:starvla} & 99.0 & 99.8 & 98.5 & 94.1 & 97.9 \\
        Qwen-VLA-Instruct~\citep{vla:qwenvla} & -- & -- & -- & -- & 97.9 \\
        HoloBrain-0-QW~\citep{vla:holobrain0} & 97.2 & 99.6 & 97.6 & 95.2 & 97.4 \\
        Pose-VLA (RGB)~\citep{vla:posevla} & 96.5 & 98.0 & 97.1 & 92.4 & 96.0 \\
        GeoPredict~\citep{vla:geopredict} & 98.0 & 98.2 & 95.7 & 94.0 & 96.5 \\
        LingBot-VA~\citep{wam:lingbot-va2026} & 98.5 & 99.6 & 97.2 & \textbf{98.5} & 98.5 \\
        ABot-M0~\citep{vla:abot-m0} & 98.8 & 99.8 & \textbf{99.0} & 96.6 & 98.6 \\
        Being-H0.7~\citep{wam:beingh07} & -- & -- & -- & -- & \textbf{99.2} \\
        DreamVLA~\citep{wam:dreamvla} & 97.5 & 94.0 & 89.5 & 89.5 & 92.6 \\
        Motus~\citep{wam:motus} & 96.8 & 99.8 & 96.6 & 97.6 & 97.7 \\
        4D-WAM~\citep{wam:4dwam} & 98.8 & 99.5 & 98.6 & 97.4 & 98.6 \\
        Fast-WAM~\citep{wam:fastwam} & 98.2 & \textbf{100.0} & 97.0 & 95.2 & 97.6 \\
        \midrule
        \textbf{WSA$_1$-B} & 98.6 & 99.6 & 97.2 & 94.2 & 97.4 \\
        \textbf{WSA$_1$-L} & \textbf{99.4} & 99.8 & 98.0 & 95.6 & 98.2 \\
        \bottomrule
    \end{tabular}
\end{table}

Table~\ref{tab:robotwin50_per_task} reports the per-task results on all 50 RoboTwin2.0 tasks.

\begin{table}[!p]
    \centering
    \caption{Per-task success rates (\%) on all 50 RoboTwin2.0 tasks. Each method is evaluated with 100 rollouts per
    task and domain. Fast-WAM and $\pi_{0.5}$ results are from
    \citet{wam:fastwam}, Appendix Table~3. Bold marks the best score
    in each domain, including ties; averages are over all 50 tasks.}
    \label{tab:robotwin50_per_task}
    \fontsize{8}{9.2}\selectfont
    \setlength{\tabcolsep}{3.2pt}
    \renewcommand{\arraystretch}{1.0}
    \begin{tabular*}{\linewidth}{@{\extracolsep{\fill}}lrrrrrrrr@{}}
        \toprule
        \textbf{Task} & \multicolumn{2}{c}{$\pi_{0.5}$}
        & \multicolumn{2}{c}{Fast-WAM}
        & \multicolumn{2}{c}{\textbf{WSA$_1$-B}}
        & \multicolumn{2}{c}{\textbf{WSA$_1$-L}} \\
        \cmidrule(lr){2-3}\cmidrule(lr){4-5}\cmidrule(lr){6-7}\cmidrule(lr){8-9}
        & Clean & Rand. & Clean & Rand. & Clean & Rand. & Clean & Rand. \\
        \midrule
        Adjust Bottle & \textbf{100} & 99 & \textbf{100} & \textbf{100} & \textbf{100} & 99 & \textbf{100} & \textbf{100} \\
        Beat Block Hammer & 96 & 93 & 99 & 97 & 98 & 95 & \textbf{100} & \textbf{100} \\
        Blocks Ranking RGB & 92 & 85 & \textbf{100} & \textbf{100} & 95 & 95 & \textbf{100} & 99 \\
        Blocks Ranking Size & 49 & 26 & \textbf{94} & \textbf{98} & 92 & 92 & 82 & 86 \\
        Click Alarmclock & 98 & 89 & \textbf{100} & \textbf{100} & \textbf{100} & \textbf{100} & 95 & 98 \\
        Click Bell & 99 & 66 & \textbf{100} & \textbf{100} & \textbf{100} & \textbf{100} & 99 & 99 \\
        Dump Bin Bigbin & 92 & 97 & \textbf{97} & 96 & \textbf{97} & \textbf{99} & \textbf{97} & 98 \\
        Grab Roller & \textbf{100} & \textbf{100} & \textbf{100} & \textbf{100} & \textbf{100} & \textbf{100} & \textbf{100} & \textbf{100} \\
        Handover Block & 66 & 57 & 95 & 81 & 91 & 84 & \textbf{96} & \textbf{85} \\
        Handover Mic & 98 & 97 & 99 & \textbf{100} & \textbf{100} & \textbf{100} & \textbf{100} & \textbf{100} \\
        \addlinespace[2pt]
        Hanging Mug & 18 & 17 & \textbf{58} & \textbf{62} & 50 & 50 & 40 & 42 \\
        Lift Pot & 96 & 85 & \textbf{100} & \textbf{100} & \textbf{100} & 99 & \textbf{100} & \textbf{100} \\
        Move Can Pot & 51 & 55 & 90 & 88 & 86 & 95 & \textbf{99} & \textbf{96} \\
        Move Pillbottle Pad & 84 & 61 & \textbf{100} & 99 & 99 & \textbf{100} & \textbf{100} & 99 \\
        Move Playingcard Away & 96 & 84 & \textbf{100} & \textbf{100} & \textbf{100} & \textbf{100} & \textbf{100} & \textbf{100} \\
        Move Stapler Pad & 56 & 42 & 77 & 64 & 65 & 75 & \textbf{82} & \textbf{81} \\
        Open Laptop & 90 & 96 & 98 & \textbf{100} & \textbf{100} & \textbf{100} & 98 & \textbf{100} \\
        Open Microwave & 34 & 77 & 62 & 45 & \textbf{100} & \textbf{100} & 90 & 77 \\
        Pick Diverse Bottles & 81 & 71 & 80 & \textbf{85} & 86 & 76 & \textbf{90} & 79 \\
        Pick Dual Bottles & 93 & 63 & \textbf{100} & \textbf{96} & 94 & 89 & 98 & \textbf{96} \\
        \addlinespace[2pt]
        Place A2B Left & 87 & 82 & 95 & 93 & 92 & \textbf{97} & \textbf{99} & \textbf{97} \\
        Place A2B Right & 87 & 84 & 93 & \textbf{99} & 94 & 94 & \textbf{98} & 95 \\
        Place Bread Basket & 77 & 64 & 91 & 93 & \textbf{99} & 94 & 98 & \textbf{97} \\
        Place Bread Skillet & 85 & 66 & 90 & \textbf{93} & 90 & 89 & \textbf{94} & 92 \\
        Place Burger Fries & 94 & 87 & 96 & 99 & \textbf{100} & \textbf{100} & \textbf{100} & \textbf{100} \\
        Place Can Basket & 62 & 62 & 71 & 69 & 80 & 79 & \textbf{81} & \textbf{80} \\
        Place Cans Plasticbox & 94 & 84 & 99 & 96 & \textbf{100} & 99 & \textbf{100} & \textbf{100} \\
        Place Container Plate & 99 & 95 & 96 & \textbf{100} & 99 & \textbf{100} & \textbf{100} & \textbf{100} \\
        Place Dual Shoes & 75 & 75 & 94 & 88 & \textbf{95} & \textbf{94} & 91 & 89 \\
        Place Empty Cup & \textbf{100} & 99 & \textbf{100} & \textbf{100} & \textbf{100} & 99 & \textbf{100} & \textbf{100} \\
        \addlinespace[2pt]
        Place Fan & 87 & 85 & 96 & 96 & 98 & \textbf{99} & \textbf{100} & 98 \\
        Place Mouse Pad & 60 & 39 & 83 & \textbf{89} & \textbf{86} & 86 & 83 & 84 \\
        Place Object Basket & 80 & 76 & 89 & 88 & 84 & 87 & \textbf{91} & \textbf{90} \\
        Place Object Scale & 86 & 80 & 90 & \textbf{97} & 85 & 88 & \textbf{96} & 93 \\
        Place Object Stand & 91 & 85 & 90 & 94 & 95 & 95 & \textbf{96} & \textbf{97} \\
        Place Phone Stand & 81 & 81 & 97 & 99 & \textbf{99} & 97 & 98 & \textbf{100} \\
        Place Shoe & 92 & 93 & 96 & 99 & 99 & 99 & \textbf{100} & \textbf{100} \\
        Press Stapler & 87 & 83 & \textbf{90} & \textbf{97} & 83 & 78 & 87 & 82 \\
        Put Bottles Dustbin & 84 & 79 & \textbf{95} & 90 & \textbf{95} & 92 & 92 & \textbf{98} \\
        Put Object Cabinet & 80 & 79 & \textbf{94} & 89 & 90 & \textbf{93} & 93 & 91 \\
        \addlinespace[2pt]
        Rotate QRcode & 89 & 87 & \textbf{93} & 89 & 87 & \textbf{92} & 88 & 90 \\
        Scan Object & 72 & 65 & 89 & 92 & 85 & 88 & \textbf{95} & \textbf{96} \\
        Shake Bottle & 99 & 97 & \textbf{100} & \textbf{100} & \textbf{100} & \textbf{100} & \textbf{100} & \textbf{100} \\
        Shake Bottle Horizontally & 99 & 99 & \textbf{100} & \textbf{100} & \textbf{100} & \textbf{100} & \textbf{100} & \textbf{100} \\
        Stack Blocks Three & 91 & 76 & 95 & 97 & 98 & 98 & \textbf{99} & \textbf{100} \\
        Stack Blocks Two & 97 & \textbf{100} & \textbf{100} & \textbf{100} & \textbf{100} & \textbf{100} & \textbf{100} & \textbf{100} \\
        Stack Bowls Three & 77 & 71 & 80 & 81 & \textbf{92} & \textbf{93} & 91 & 91 \\
        Stack Bowls Two & 95 & 96 & 92 & 98 & 97 & \textbf{100} & \textbf{99} & \textbf{100} \\
        Stamp Seal & 79 & 55 & \textbf{90} & \textbf{94} & 73 & 82 & 75 & 89 \\
        Turn Switch & 62 & 54 & 61 & 59 & 61 & \textbf{75} & \textbf{66} & 73 \\
        \midrule
        \textbf{Average} & 82.74 & 76.76 & 91.88 & 91.78 & 92.18 & 92.70 & \textbf{93.52} & \textbf{93.14} \\
        \bottomrule
    \end{tabular*}
\end{table}

\Needspace{5\baselineskip}
\subsection{Robustness under scene-level shift}
Table~\ref{tab:clean_to_random_shifts} isolates the effect of state and action normalization on transfer to the unseen ARX-5 robot. One WSA$_1$-B checkpoint is evaluated twice, first with the ALOHA means and standard deviations supplied with the checkpoint and then with 14-dimensional statistics recomputed from the official RoboTwin ARX-5 demonstrations on Hugging Face. These statistics standardize the input state and convert normalized action predictions back to their original scale, while the policy parameters, the embodiment mapping and the evaluation settings are held fixed. Switching the statistics raises success from 17.0\% to 47.1\% in the clean setting and from 17.2\% to 47.6\% under randomization, differences of 30.1 and 30.4 points. The absolute level of a zero-shot cross-embodiment result is therefore a property of the normalization statistics as much as of the policy, and the WSA$_1$-B entry of Table~\ref{tab:ood-ce} is the one obtained with ARX-5 statistics.

\begin{table}[!htbp]
    \centering
    \caption{Performance drop under OOD-C2R setting. $\Delta$ is Randomized$-$Clean in percentage points.}
    \label{tab:clean_to_random_shifts}
    \small
    \setlength{\tabcolsep}{5pt}
    \renewcommand{\arraystretch}{1.08}
    \begin{tabular}{lrrr}
        \toprule
        \textbf{Task} & \textbf{Clean} & \textbf{Rand.} & $\boldsymbol{\Delta}$ \\
        \midrule
        Open Microwave & 71 & 32 & $-39$ \\
        Click Bell & 95 & 65 & $-30$ \\
        Handover Block & 74 & 46 & $-28$ \\
        Put Object Cabinet & 56 & 36 & $-20$ \\
        Beat Block Hammer & 94 & 75 & $-19$ \\
        \midrule
        Place Can Basket & 52 & 64 & $+12$ \\
        Turn Switch & 49 & 53 & $+4$ \\
        Pick Diverse Bottles & 75 & 79 & $+4$ \\
        Hanging Mug & 29 & 33 & $+4$ \\
        \bottomrule
    \end{tabular}
\end{table}

\Needspace{5\baselineskip}
\subsection{Cross-embodiment transfer and normalization statistics}
Table~\ref{tab:arx5_normalization} examines how state and action normalization
affects transfer to ARX-X5. We compare the ALOHA means and standard deviations
supplied with the WSA$_1$-B checkpoint with 14-dimensional statistics estimated
from the official RoboTwin ARX-X5 demonstrations on Hugging Face. These
statistics standardize the input state and convert normalized action
predictions back to their original scale. Policy parameters, embodiment
mapping, and evaluation settings remain fixed. Using ARX-X5 statistics
improves success by 30.1 percentage points in Clean and 30.4 points in
Randomized, showing that normalization substantially affects transfer with a
fixed policy. The WSA$_1$-B results in Table~\ref{tab:ood-ce} use ARX-X5
statistics.
\begin{table}[!htbp]
    \centering
    \caption{Effect of normalization statistics on WSA$_1$-B transfer to
    ARX-X5. Success rates (\%) are measured over 45 tasks with 100 episodes
    per task and setting, using the same checkpoint.}
    \label{tab:arx5_normalization}
    \small
    \begin{tabular}{lcc}
        \toprule
        \textbf{Statistics source} & \textbf{Clean} & \textbf{Randomized} \\
        \midrule
        ALOHA (checkpoint) & 17.0 & 17.2 \\
        ARX-X5 (recomputed) & 47.1 & 47.6 \\
        \bottomrule
    \end{tabular}
\end{table}

\Needspace{5\baselineskip}
\subsection{More details of WSA modeling}
\label{app:additional_ablations}

\paragraph{World--action modeling directions.}
Table~\ref{tab:mask_implementation_map} records which information paths each setting of the directional ablation retains, and corresponds row by row to the four settings compared in Table 3. Removing G$\rightarrow$A blocks scene information from both the state and the action representations, including the indirect scene$\rightarrow$state$\rightarrow$action path, and costs 9.2 points of average success; the fully unidirectional baseline removes G$\rightarrow$V as well and uses triangular attention, yet its success rate matches the variant that removes action-to-world alone, which is why Section 4.1 attributes the gain to the coupling between the two directions rather than to the number of prediction targets. Figure 4 shows the same ablation qualitatively, plotting future point clouds decoded with frozen DA3 heads for four WSA$_1$-B variants alongside the teacher reference. The bidirectional model follows the reference geometry most closely and the ablated variants depart from it, so the ordering of the columns follows the ordering of success rates in Table 3, which is consistent with the reading that a more accurate learned world accompanies stronger manipulation generalization.

\begin{table}[!htbp]
    \centering
    \caption{Information-flow paths in the directional ablation.}
    \label{tab:mask_implementation_map}
    \small
    \begin{tabular*}{\linewidth}{@{\extracolsep{\fill}}lccc@{}}
        \toprule
        \textbf{Setting} & \textbf{G$\rightarrow$V}
        & \textbf{A$\rightarrow$G} & \textbf{G$\rightarrow$A} \\
        \midrule
        Fully unidirectional & No & No & Yes \\
        Fully bidirectional & Yes & Yes & Yes \\
        w/o action-to-world & Yes & No & Yes \\
        w/o world-to-action & Yes & Yes & No \\
        \bottomrule
    \end{tabular*}
\end{table}

\Needspace{5\baselineskip}
\subsection{Visualization of predicted 3D world}
\begin{figure}[!htbp]
    \centering
    \includegraphics[width=0.90\linewidth]{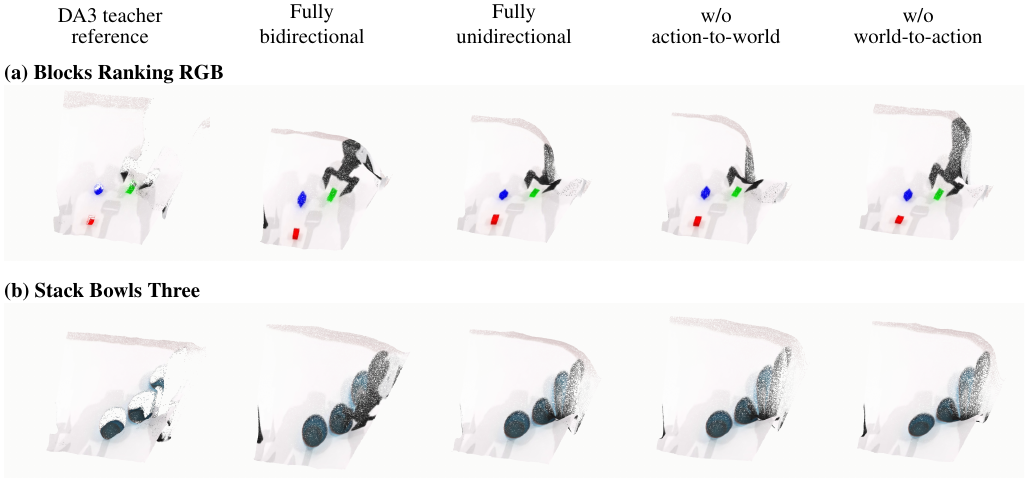}
    \caption{Decoded future point clouds for the directional attention ablation
    on (a) Blocks Ranking RGB and (b) Stack Bowls Three. The first column shows
    the DA3 teacher reference from the future observation. The remaining columns
    show point clouds obtained by decoding the predicted 3D features with
    frozen DA3 heads.}
    \label{fig:pointcloud_directional_ablation}
\end{figure}

We compare the predicted point clouds across WSA\(_1\)-B variants. As shown in Figure~\ref{fig:pointcloud_directional_ablation}, bidirectional WSA modeling yields the best performance, indicating that the better the 3D world representation the model learns, the stronger manipulation generalization it achieves.

\Needspace{4\baselineskip}
\subsection{Real-Robot Tasks and Per-Task Results}
\label{app:real_robot_details}

We use two platforms: a 7-DoF single-arm AgileX PiPER and a 14-DoF bimanual
ARX Lift2, each observed by three cameras. Trials begin from predefined object
positions, orientations, and layouts. Success requires completion of the full
task. Completeness is the fraction of predefined task stages completed before
termination. Trial counts and stage-progress rules are specific to each task,
and task-level scores are averaged with equal weight. We train and evaluate
all five policies in Table~\ref{tab:real_robot_results} using the same downstream
demonstrations and evaluation settings across the seven tasks. Table~\ref{tab:real_robot_setup} summarizes the tasks and demonstration data.

The evaluation sets contain 30 trials for each of the first five tasks,
34 trials for Sweep Trash, and 20 trials for Unzip Pencil Bag. In the order
listed in the table, WSA$_1$-L completes the full task in 25/30, 23/30, 26/30,
23/30, 21/30, 29/34, and 17/20 trials. Averaging the seven task success rates
with equal weight gives 80.52\%.

\begin{table}[!htbp]
    \centering
    \caption{Per-task real-robot results (\%). SR denotes full-task success and
    C denotes stage completeness. Each policy uses the same number of trials
    per task. Averages are unweighted means over the seven tasks, computed
    separately for SR and C.}
    \label{tab:real_robot_results}

    \small
    \setlength{\tabcolsep}{3.5pt}
    \renewcommand{\arraystretch}{1.08}

    \resizebox{\textwidth}{!}{%
        \begin{tabular}{lc|cc|cc|cc|cc|cc}
            \toprule
            \multirow{2}{*}{\textbf{Task}}
            & \multirow{2}{*}{\textbf{Robot}}
            & \multicolumn{2}{c|}{$\pi_0$}
            & \multicolumn{2}{c|}{$\pi_{0.5}$}
            & \multicolumn{2}{c|}{InternVLA-A1}
            & \multicolumn{2}{c|}{WSA$_1$-B}
            & \multicolumn{2}{c}{WSA$_1$-L} \\

            & & SR & C & SR & C & SR & C & SR & C & SR & C \\
            \midrule

            RGB Block Sorting
            & PiPER
            & 70.00 & 70.00
            & 70.00 & 70.00
            & 50.00 & 50.00
            & 80.00 & 80.00
            & \textbf{83.33} & \textbf{83.33} \\

            Object Organization
            & PiPER
            & 50.00 & 50.00
            & 76.67 & 76.67
            & 36.67 & 36.67
            & \textbf{80.00} & \textbf{80.00}
            & 76.67 & 76.67 \\

            \midrule

            Trash Cleaning
            & Lift2
            & 36.67 & 53.33
            & 76.67 & 86.67
            & 43.33 & 58.33
            & 80.00 & 90.00
            & \textbf{86.67} & \textbf{91.67} \\

            Pen Holder Placement
            & Lift2
            & 6.67 & 28.33
            & 70.00 & 76.67
            & 33.33 & 38.33
            & \textbf{86.67} & \textbf{90.00}
            & 76.67 & 76.67 \\

            Toy Box Organization
            & Lift2
            & 13.33 & 18.33
            & 13.33 & 26.67
            & 36.67 & 48.33
            & 60.00 & 71.67
            & \textbf{70.00} & \textbf{85.00} \\

            Sweep Trash
            & Lift2
            & 29.41 & 40.19
            & 26.47 & 44.12
            & 44.12 & 78.41
            & 79.41 & 89.21
            & \textbf{85.29} & \textbf{93.12} \\

            Unzip Pencil Bag
            & Lift2
            & 30.00 & 42.50
            & 55.00 & 62.50
            & 25.00 & 42.50
            & 75.00 & 75.00
            & \textbf{85.00} & \textbf{90.00} \\

            \midrule

            \multicolumn{2}{c|}{\textit{Average}}
            & 33.73 & 43.24
            & 55.45 & 63.33
            & 38.45 & 50.37
            & 77.30 & 82.27
            & \textbf{80.52} & \textbf{85.21} \\

            \bottomrule
        \end{tabular}
    }
\end{table}

\begin{table}[!htbp]
    \centering
    \caption{Real-robot tasks and demonstration data. Traj. / Time denotes
    the trajectory count and total collection time.}
    \label{tab:real_robot_setup}

    \small
    \setlength{\tabcolsep}{3pt}
    \renewcommand{\arraystretch}{1.08}

    \resizebox{\textwidth}{!}{%
        \begin{tabular}{
            >{\centering\arraybackslash}m{0.29\textwidth}
            |>{\centering\arraybackslash}m{0.16\textwidth}
            |>{\centering\arraybackslash}m{0.15\textwidth}
            |>{\arraybackslash}m{0.40\textwidth}
        }
            \toprule
            \textbf{Initial / Final}
            & \textbf{Robot}
            & \textbf{Traj. / Time}
            & \textbf{Task} \\
            \midrule

            \includegraphics[width=0.28\textwidth]{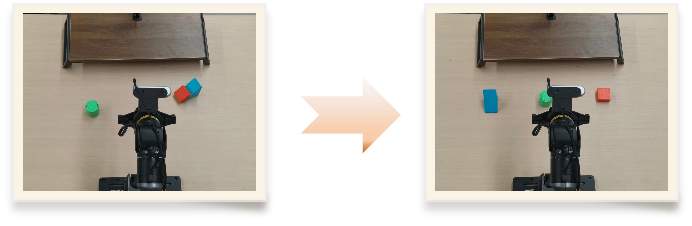}
            & \shortstack{AgileX PiPER\\7-DoF}
            & 30 / 21 min
            & \textbf{RGB Block Sorting}: place red, green, and blue blocks
            from left to right in the specified order. \\
            \midrule

            \includegraphics[width=0.28\textwidth]{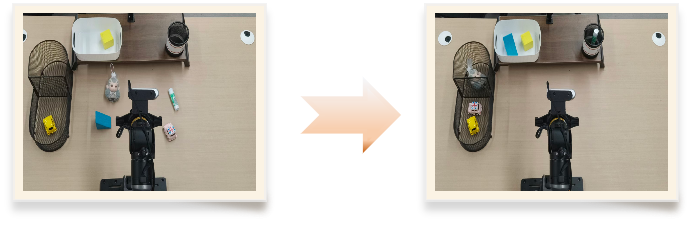}
            & \shortstack{AgileX PiPER\\7-DoF}
            & 50 / 39 min
            & \textbf{Object Organization}: sort toys, stationery, and trash
            into their designated containers. \\
            \midrule

            \includegraphics[width=0.28\textwidth]{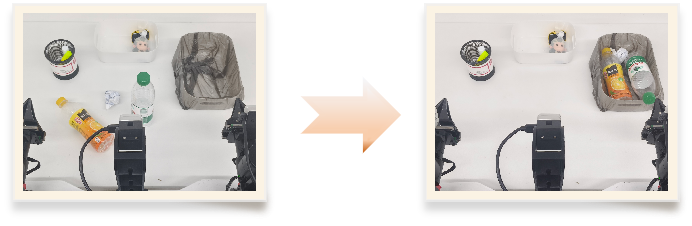}
            & \shortstack{ARX Lift2\\14-DoF}
            & 30 / 5 min
            & \textbf{Trash Cleaning}: grasp trash, hand it between grippers,
            and place it in the bin. \\
            \midrule

            \includegraphics[width=0.28\textwidth]{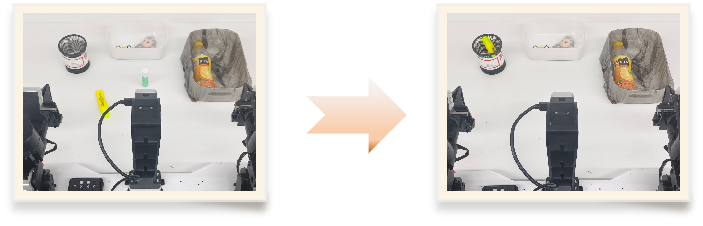}
            & \shortstack{ARX Lift2\\14-DoF}
            & 30 / 4 min
            & \textbf{Pen Holder Placement}: pick up stationery and place it
            in the pen holder. \\
            \midrule

            \includegraphics[width=0.28\textwidth]{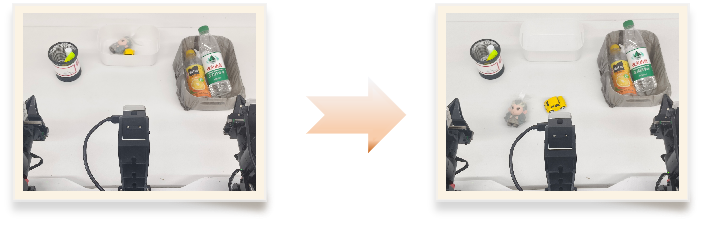}
            & \shortstack{ARX Lift2\\14-DoF}
            & 30 / 4 min
            & \textbf{Toy Box Organization}: place tabletop toys into the toy
            box and arrange them. \\
            \midrule

            \includegraphics[width=0.28\textwidth]{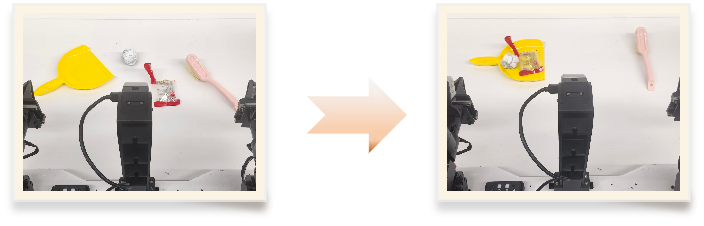}
            & \shortstack{ARX Lift2\\14-DoF}
            & 90 / 31 min
            & \textbf{Sweep Trash}: use a broom to sweep trash into a dustpan. \\
            \midrule

            \includegraphics[width=0.28\textwidth]{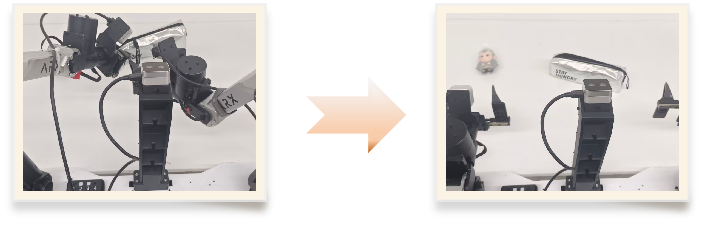}
            & \shortstack{ARX Lift2\\14-DoF}
            & 30 / 8 min
            & \textbf{Unzip Pencil Bag}: stabilize the bag, unzip it, and
            return it to the tabletop. \\

            \bottomrule
        \end{tabular}
    }
\end{table}

\clearpage
\subsection{World and Action Predictions}
\label{app:world_action_visualizations}
\begin{figure}[!htbp]
    \centering
    \includegraphics[width=0.95\linewidth,height=0.75\textheight,keepaspectratio]{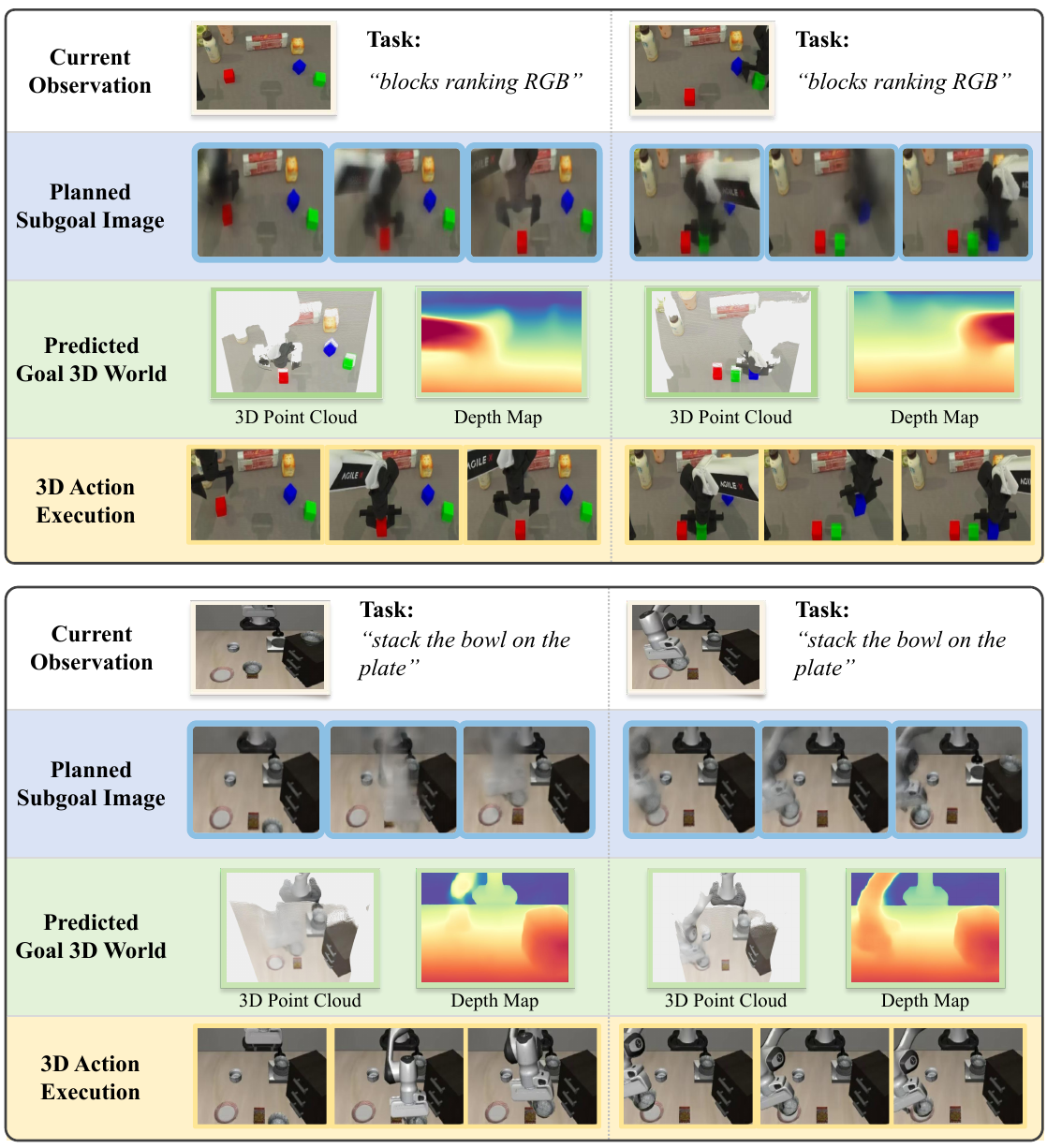}
    \caption{Representative WSA$_1$ predictions. Rows show the current
    observation and instruction, predicted subgoal image, depth and point-cloud
    visualizations obtained by decoding predicted 3D features with frozen
    DA3 heads, and action execution.}
    \label{fig:vis_study}
\end{figure}

To reveal how WSA$_1$ performs robot control, we present qualitative visualizations of the model’s predictions on various tasks. As illustrated in Figure~\ref{fig:vis_study}, the visualization covers four views: visual observation with text instruction, predicted future images, goal 3D world prediction with 3D point cloud and depth map, and trajectory of executed actions.
From the results, we find that predicted visual plans well capture the task objective and well align the generated 3D world. Furthermore, driven by explicit 3D world modeling, the action trajectories are well-aligned with the predicted 3D goal states. 
These visual results demonstrate that WSA$_1$ unifies visual thinking, 3D world dynamics, and physical action generation within a shared latent space. It jointly models how actions alter the 3D world and how 3D world states guide action generation, thus leading to robust and generalizable manipulation behaviors.

\clearpage
\subsection{Cross-Embodiment Execution}
\label{app:cross_embodiment_visualizations}
\begin{figure}[!htbp]
    \centering
    \includegraphics[width=\linewidth,height=0.75\textheight,keepaspectratio]{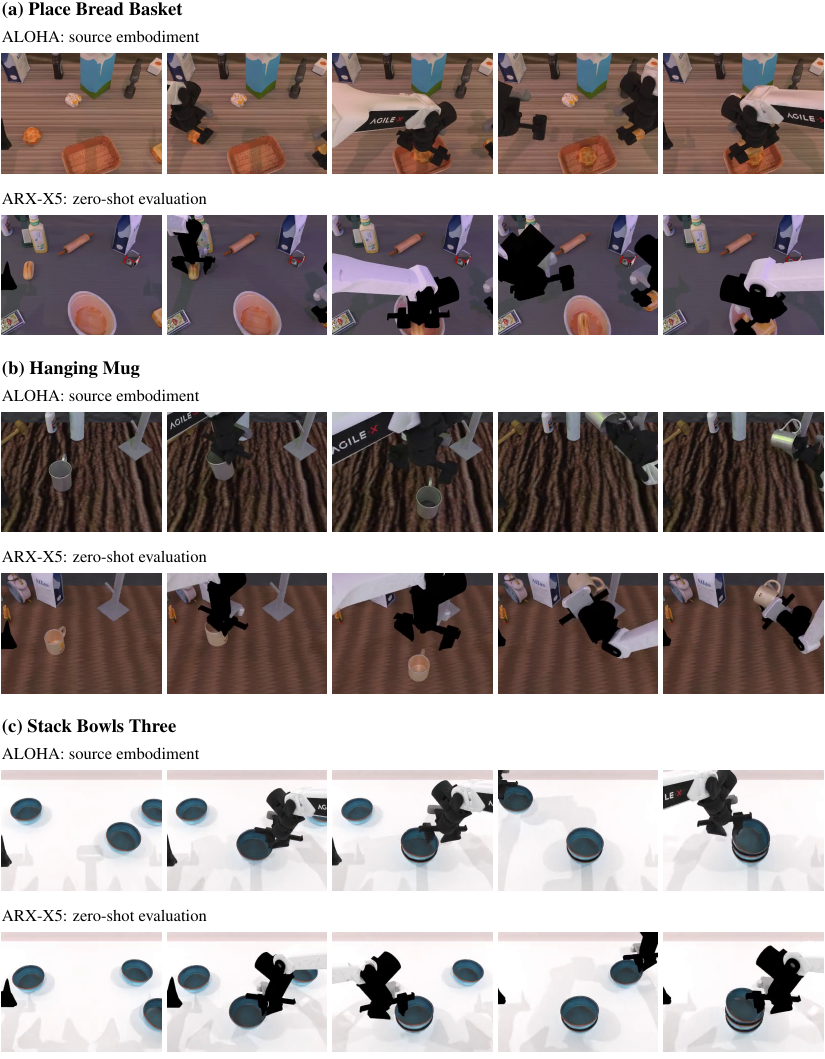}
    \caption{Cross-embodiment execution on three RoboTwin tasks:
    (a) \textit{Place Bread Basket}, (b) \textit{Hanging Mug}, and
    (c) \textit{Stack Bowls Three}. Each task shows an ALOHA rollout and an
    ARX-X5 zero-shot rollout, with five frames progressing from left to right
    within each row. The ARX-X5 rollouts use a policy adapted on ALOHA
    demonstrations without target-embodiment policy fine-tuning.}
    \label{fig:robotwin_embodiment_rollouts}
\end{figure}

Figure~\ref{fig:robotwin_embodiment_rollouts} shows execution sequences
for three RoboTwin tasks on the source ALOHA embodiment and on ARX-X5
under zero-shot transfer.

\clearpage
\subsection{Long-Horizon Real-Robot Execution}
\label{app:long_horizon_visualizations}
\begin{figure}[!htbp]
    \centering
    \includegraphics[width=\linewidth,height=0.75\textheight,keepaspectratio]{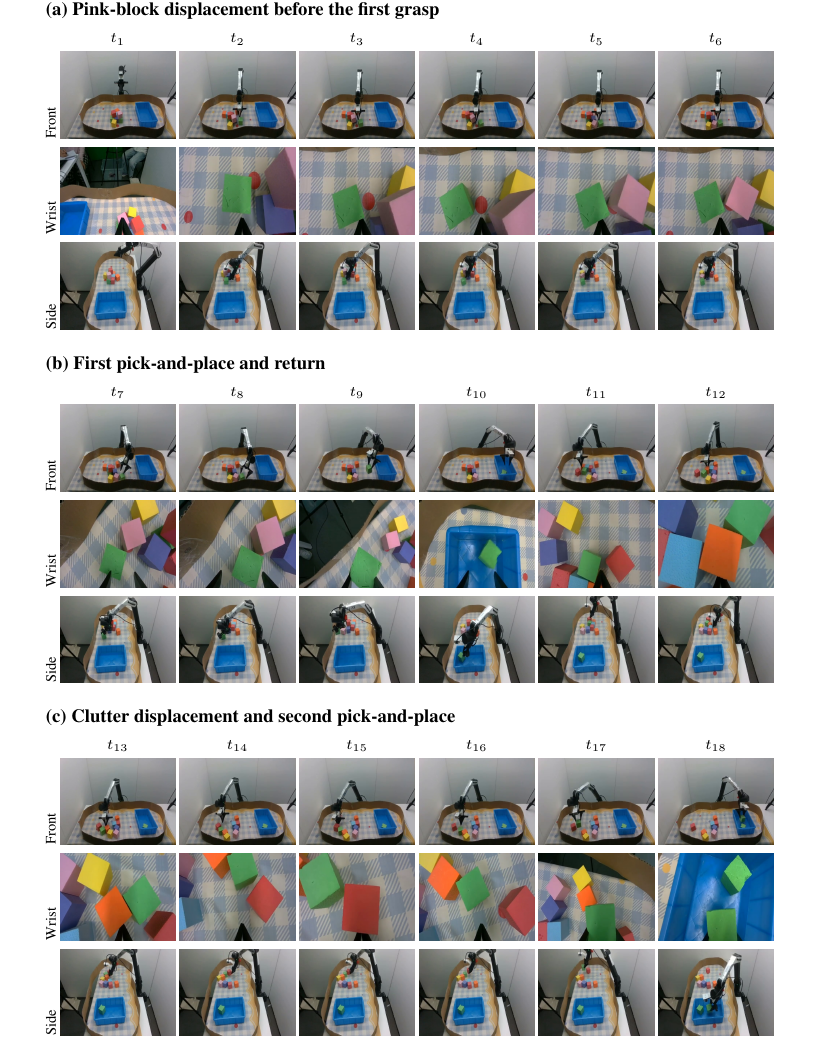}
    \caption{Long-horizon WSA$_1$-B execution of \textit{Pick Out the Green
    Blocks} on RoboChallenge 2.0. Eighteen selected time points ($t_1$--$t_{18}$)
    progress from left to right within panels (a)--(c). Each time point is
    shown from synchronized front, wrist, and side views. The sequence
    captures the pink block being pushed aside before the first grasp,
    further displacement of non-target blocks, and two successive transfers
    of green blocks from the cluttered pile to the blue container.}
    \label{fig:robochallenge_green_blocks}
\end{figure}

Figure~\ref{fig:robochallenge_green_blocks} shows WSA$_1$-B performing
\textit{Pick Out the Green Blocks} on RoboChallenge 2.0
(Table30 V2)~\citep{data:robochallenge_v2}.

\end{document}